\documentclass[11pt]{article}

\usepackage[final]{acl}

\usepackage{times}
\usepackage{latexsym}

\usepackage[T1]{fontenc}

\usepackage[utf8]{inputenc}

\usepackage{microtype}

\usepackage{inconsolata}

\usepackage{graphicx}

\usepackage{subcaption}
\usepackage{algorithmic}
\usepackage{algorithm}
\usepackage{amsmath}
\usepackage{graphicx} 
\usepackage{multirow} 
\usepackage{hyperref}%
\usepackage{url}%
\usepackage{booktabs}%
\usepackage{amsfonts}%
\usepackage{nicefrac}%
\usepackage{microtype}%
\usepackage{xcolor}%
\usepackage{booktabs}
\usepackage{adjustbox}  
\usepackage{listings}
\usepackage{xcolor}
\usepackage{longtable}
\usepackage{array}
\usepackage{ragged2e}
\usepackage{pdflscape}
\usepackage{centernot}
\usepackage{amssymb}

\lstdefinestyle{prompt}{
    basicstyle=\ttfamily\scriptsize,
    breaklines=true,
    breakatwhitespace=true,
    columns=fullflexible,
    keepspaces=true,
    frame=single,
    framesep=3pt,
    xleftmargin=2pt,
    xrightmargin=2pt,
    aboveskip=4pt,
    belowskip=4pt,
    showstringspaces=false
  }
  
\title{When Objectives and Evaluators Disagree: Evaluating Transfer for Post-hoc Demographic Framing Mitigation}

\author{Tunazzina Islam\thanks{Now at the University of Southern California (\texttt{tunazzin@usc.edu}).} \\
    Department of Computer Science \\
    Purdue University, West Lafayette \\
    Indiana - 47907, USA \\
    \texttt{islam32@purdue.edu}}
    
\begin{document}
\maketitle
\begin{abstract}

Post-hoc rewriting is used to reduce demographic disparities in LLM-generated text when the source model is inaccessible. A rewrite can look successful merely because it improved its own target or satisfied its selector. We study this as a transfer problem on 660 gender pairs and 1,980 age contrasts of demographic-conditioned climate messages. A rewriter proposes edits, a selection model chooses among constraint-satisfying candidates, and an independent evaluator uninvolved in either rates pairwise framing disparity. A fixed policy that always targets the message with the lower Persuasion Bias Index (PBI) transfers inconsistently across axes and outcome aggregations. An adaptive policy that proposes edits on both sides but accepts at most one significantly improves on that rule, applied to the same candidates, in seven of eight comparisons across two evaluator families, mostly by editing more pairs, not by guided direction choice, at higher preservation risk. A policy editing both messages gives the largest point reductions on fewer pairs. These two policies reduce disparity across a preservation-threshold sweep; absolute ratings vary across evaluator families and prompts. Shrinking the PBI gap increases dimension-mean disparity under both evaluator families. Mitigation should be reported as transfer across objectives, evaluators and their prompts, and editing policies.

\end{abstract}

\section{Introduction}

Large language models (LLMs) \citep{brown2020language} are increasingly used to generate text tailored to specific demographic groups, enabling personalized communication in domains such as policy outreach, marketing, and public messaging \citep{liu2025llms,hackenburg2024evaluating,breum2024persuasive}. While such demographic conditioning can improve relevance and engagement, it also raises fairness concerns: models may systematically vary not only the content of messages, but also their tone, framing, and persuasive style across demographic groups \citep{islam2026gets,pauli2026analysing,pauli2025measuring,tan2025unmasking,fang2024bias,wan2023kelly}. These differences motivate methods for auditing and mitigating demographic disparities in generated text.

Many practical systems provide access only to completed outputs rather than model parameters, training data, or decoding procedures. In such settings, mitigation must be performed \textit{post hoc} through rewriting, filtering, or selective editing \citep{wan2025white,banerjee2024all,stahl2022prefer,krause2021gedi,schick2021self,dixon2018measuring}. A common procedure is to define a disparity score, generate a revised output, and retain the revision when that score improves. However, this creates an evaluation ambiguity.
A rewrite may appear successful because it genuinely improves a separately measured outcome, or because the same objective or evaluator influenced both candidate selection and final evaluation.

\begin{figure*}
\includegraphics[width=\textwidth]{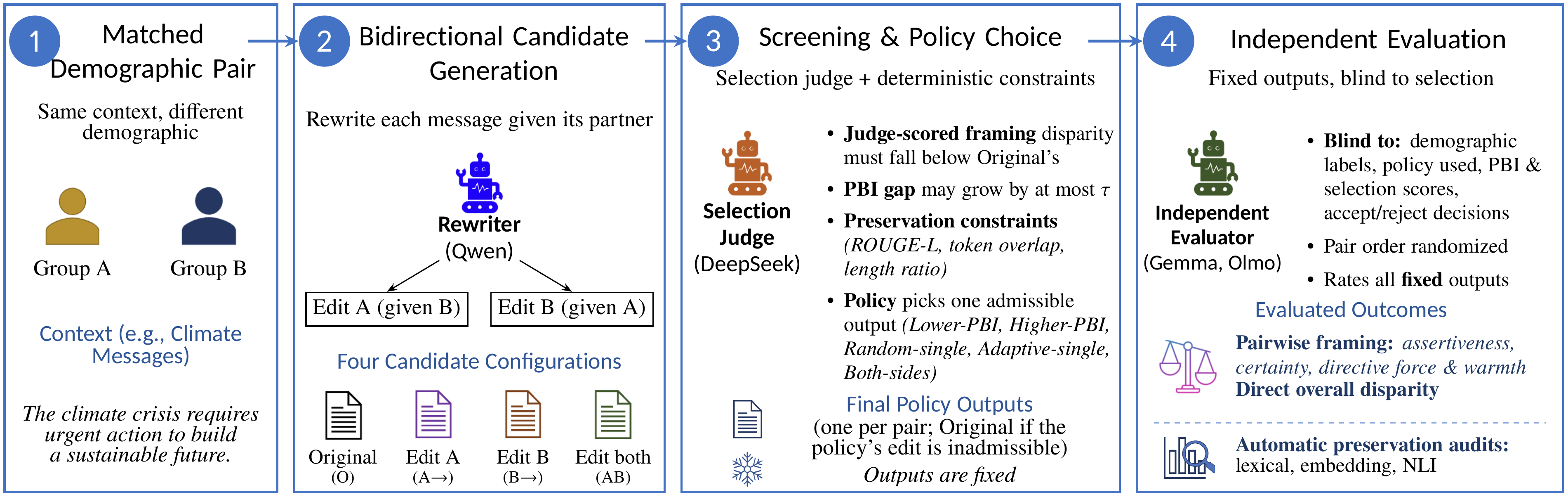}
\caption{{\small Overview of the framework. (1) Matched pairs differ only in the
demographic audience. (2) A rewriter proposes an edit of each message given its
partner. (3) A selection judge and deterministic PBI and preservation
constraints decide which edits are admissible, and an editing policy fixes one
output per pair, falling back to Original when its edit is inadmissible.
(4) A separate evaluator, blind to all selection information, rates the fixed
outputs.}}
\vspace{-5pt}
\label{fig:frame}
\vspace{-5pt}
\end{figure*}
Our pipeline separates selection from evaluation (Fig.~\ref{fig:frame}).  A rewriting model proposes edits to matched pairs of messages written for two demographic audiences; a \textbf{selection judge}, together with deterministic tolerance and text-preservation constraints, decides which edits are accepted; and after every final output is fixed, an \textbf{independent evaluator}, which took no part in generation or selection, rates the finalized pairs. No model occupies both
roles, and every policy effect we estimate is a paired difference between two
conditions of the same matched pair, scored by one evaluator within a single run. We formulate this attribution question as a transfer problem spanning objectives, evaluators, and editing policies. \textbf{Objective transfer} asks whether improving one operational measure of disparity also improves another. For example, reducing a deterministic persuasion gap based on agency, certainty, and imperative language may not reduce pairwise differences in warmth or overall communication style. \textbf{Evaluator transfer} asks whether an improvement measured by the evaluator that selected a rewrite remains visible to an evaluator that did not participate in candidate generation or acceptance, including under a reworded evaluation prompt. When the same LLM judgment is used for both selection and final reporting, improvement on that score is partly guaranteed by construction and cannot independently establish effectiveness.

We study these questions using matched demographic pairs of climate change-related messages generated by three source models \citep{islam2026gets}. Each pair holds the source model, stance, theme, region, and other content variables fixed while varying gender or age. The dataset contains 660 gender pairs and 1,980 age contrasts drawn from 330 four-age-group contexts. This matched design isolates text differences associated with the demographic condition under the observed generation setup, but it is a diagnostic comparison rather than a claim of causal counterfactual fairness.

We evaluate \textbf{three} operational views of disparity. The first is the Persuasion Bias Index (PBI) introduced by \citet{islam2026gets}, a \textit{deterministic message-level score} based on agency, certainty, and imperative features. The other two are \textit{pairwise LLM-judged outcomes}: an equal-weight mean over assertiveness, certainty, directive force, and warmth, and a separate direct rating of overall framing disparity. These operationalizations partially overlap. In particular, assertiveness, certainty, and directive force are closely related to PBI components, while warmth provides the clearest extension. We therefore do not treat them as independent constructs or validated definitions of fairness. Instead, we ask whether conclusions transfer across their different units of analysis, elicitation procedures, and aggregation rules.

The independent evaluator receives no demographic, policy, PBI, acceptance, or selection-score metadata; this does not make its judgments objective, but it prevents the evaluator behind the final claims from determining which outputs
enter the comparison.

Our initial analysis considers a \textit{fixed-direction policy} that always rewrites the message with lower original PBI. Its improvements transfer inconsistently across demographic axes and outcome definitions, suggesting that edit direction itself is an important policy choice. We therefore propose \textbf{Adaptive-single}, a pair-aware method that generates both one-sided edit directions and selects at most one admissible candidate. For controlled comparison, all direction policies are constructed from the same candidate pool and evaluated only after their outputs are finalized.

Under independent evaluation, \textbf{Adaptive-single} outperforms the controlled Lower-PBI policy on both outcomes for age and on the dimension mean for gender. A Both-sides policy, which changes both messages when accepted, gives the largest point reductions on both outcomes while editing fewer pairs, but is separable from Adaptive-single only for age under a second evaluator family. Preservation audits further show that greater mitigation coverage can increase policy-level exposure to possible content drift, even when conditional embedding and natural language inference (NLI) diagnostics are similar across policies.
These findings motivate evaluating post-hoc mitigation jointly in terms of outcome transfer, evaluator separation, edit direction, and intervention risk.

Our contributions are\footnote{All codes and data are available here: \href{https://github.com/tunazislam/posthoc-mitigation-transfer-evaluation}{https://github.com/tunazislam/posthoc-mitigation-transfer-evaluation} }:
\begin{itemize}\setlength{\itemsep}{0pt}\setlength{\parskip}{0pt}
    \item We formulate post-hoc demographic mitigation as transfer across
    operational objectives, evaluators and editing policies, and introduce an
    independent-evaluation protocol that separates candidate selection from
    final assessment while keeping unchanged fallbacks in the estimand.
    \item We introduce Adaptive-single, a pair-aware policy that evaluates both
    one-sided edit directions, and show that its advantage over fixed Lower-PBI
    editing comes mostly from wider intervention coverage, with guided direction
    choice adding a smaller but significant share on the dimension mean.
    \item We show that the optimized objective and the independently measured
    outcome can disagree in \emph{sign}, not only in magnitude, under two
    evaluator families.
    \item We test our conclusions against a second evaluator family, prompt
    variants for the fixed policy and a threshold sweep, and report
    preservation, qualitative and exploratory human diagnostics showing
    lower disparity scores coexisting with evaluator disagreement or content
    risk.
\end{itemize}

We study three questions: whether selection-time improvements transfer to an independent evaluator (RQ1), whether adaptive edit-direction selection outperforms fixed Lower-PBI editing (RQ2), and whether PBI-targeted improvements transfer to pairwise framing outcomes (RQ3). 

\section{Related Work}
\paragraph{Bias and fairness in LLM generation.}
Demographic conditioning can make LLM outputs differ systematically in content,
tone and framing across groups
\citep{wang2026invests,islam2026gets,pauli2026analysing,pauli2025measuring,sethi2025ai,tan2025unmasking,aghaebe2025llms,fang2024bias,dai2024bias,kumar2023language,li2023fairness,li2023survey,sag2023fairness,urman2023silence,kotek2023gender,esiobu2023robbie,wan2023kelly},
including in lexical choice, stylistic framing and persuasive intensity
\citep{islam2026gets,wan2023kelly}, and such biases depend on the demographic
distribution of training data \citep{garimella2022demographic}.

\paragraph{Bias mitigation and post-hoc rewriting.}
Bias is mitigated through data augmentation, model adaptation, controlled
decoding, filtering, output aggregation and reasoning-step intervention
\citep{abels2025wisdom,wu-etal-2025-reasoning,banerjee2024all,masoudian-etal-2024-effective,krause2021gedi,schick2021self,dinan2020queens},
which typically require access to training data, model internals or
intermediate reasoning. Post-hoc methods instead rewrite finished outputs, for
example by modifying agency and power or revising biased spans
\citep{wan2025white,stahl2022prefer}. Unlike these single-output approaches, we
model disparity over matched demographic pairs and separate edit coverage from
mitigation effect, and lexical similarity from semantic preservation.

\paragraph{Pairwise evaluation and competing objectives.}
Matched demographic comparisons hold the content context fixed while varying
the demographic condition
\citep{zhou2024counterfactual,banerjee2024all,rosenblatt2023counterfactual,kusner2017counterfactual},
but different fairness criteria can conflict, and optimizing one need not
improve another \citep{chouldechova2017fair,kleinberg2016inherent}. We study
this in generative mitigation by comparing a deterministic message-level
persuasion objective with pairwise framing judgments.

\paragraph{LLM judges and evaluator circularity.}
LLMs are widely used as evaluators
\citep{brady-islam-2026-iterative,sudhoff2026paid,islam2026reasoning,tan2025judgebench,islam2025uncovering,islam2025discovering,zheng2023judging,liu2023g,chiang2023can,li2024llms,lin2022truthfulqa},
yet their judgments are sensitive to prompt, model, position and scale anchors
\citep{shi2025judging,sclar2024quantifying,zhuo2024prosa}. Using one judge
score for both acceptance and final evaluation is circular even with a separate
rewriter, so we separate the selection judge from an independent evaluator and
treat their disagreement as an empirical transfer result.

\paragraph{Transfer of mitigation effects and of evaluators.}
\citet{jin2021transferability} ask whether fine-tuned bias-mitigation gains
transfer across tasks and domains, largely holding the evaluation construct
fixed; CAFIE \cite{banerjee2024all} intervenes at decoding time so that
counterfactual continuations are treated alike; and JudgeBench
\cite{tan2025judgebench} tests LLM judges in settings where a verifiable answer
adjudicates their verdicts. We instead hold the model and data fixed and vary
the \emph{measurement}: the objective a post-hoc rewriter optimizes, the prompt
that elicits the outcome, and the evaluator family that scores it. With no
validated ground-truth fairness quantity to adjudicate between these
operationalizations, their disagreement must be reported rather than resolved,
and we show that it can extend to the sign of the effect.

\section{Problem Formulation}
\label{sec:problem_formulation}

\subsection{Matched Demographic Outputs}
\label{sec:matched_outputs}

Let $x \in \mathcal{X}$ denote the non-demographic message context, including the source model, stance, theme, region, and other fixed content variables. Let $a \in \mathcal{A}$ denote a demographic condition. A fixed black-box generator $G$ produces $y = G(x,a)$. For instance $i$, we construct a matched pair
\[
Y_i
=
\left(y_i^{A},y_i^{B}\right)
=
\left(
G(x_i,a_i^{A}),
G(x_i,a_i^{B})
\right)
\]
where $x_i$ is held fixed and $a_i^{A} \neq a_i^{B}$ varies within one demographic axis.
For gender, each fixed context produces one male--female contrast. For age, each underlying context contains four messages and contributes all six unordered pairwise contrasts. Therefore, age-pair observations derived from the same four-message context are statistically dependent and are clustered at the underlying-context level during inference.

This matched design isolates observed differences associated with changing the demographic condition while holding the recorded generation context fixed. It is an auditing design and does not, by itself, establish counterfactual fairness in the causal sense.

\subsection{Message-Level Persuasion Disparity}
\label{sec:pbi_formulation}

Following \citet{islam2026gets}, let $P(y)$ denote the Persuasion Bias Index (PBI):
\[
P(y)
=
A(y) + C(y) + \lambda I(y)
\]
where $A(y)$ is a lexicon-based agency score, $C(y)$ captures certainty and hedging, $I(y)$ captures imperative force, and $\lambda = 0.1$. For a matched pair $Y_i$, the absolute PBI disparity is
\[
d_P(Y_i)
=
\left|
P(y_i^{A}) - P(y_i^{B})
\right|
\]
PBI is deterministic for fixed text. We treat it as an operational linguistic score, not a full measure of persuasion, demographic disparity or fairness.
\subsection{Pairwise Framing Outcomes}
\label{sec:framing_outcomes}

For a message pair $Y_i$, an LLM evaluator $J$ assigns ordinal disparity ratings for each framing dimension $d_{J,k}(Y_i) \in \{0,1,2,3\}$ where 
$k \in
\{
\text{assertiveness},
\text{certainty},
\text{directive},
\text{warmth}
\}$.
We define the equal-weight dimension outcome as
\[
d_{J,\mathrm{dim}}(Y_i)
=
\frac{1}{4}
\sum_{k=1}^{4}
d_{J,k}(Y_i)
\]
The evaluator separately provides a direct overall-disparity rating
$d_{J,\mathrm{overall}}(Y_i)
\in
\{0,1,2,3\}$, which is elicited directly and is not computed from the four-dimensional ratings.

These outcomes are operational LLM judgments. Their dimensions partially overlap conceptually with PBI: certainty corresponds directly to the PBI certainty component, directive force is closely related to imperative force, and assertiveness may overlap with agency. Warmth provides the clearest extension beyond PBI. We therefore do not treat PBI and the LLM-judged outcomes as independent constructs or as validated definitions of fairness. Instead, we study whether conclusions transfer across their different units of analysis, elicitation procedures, and aggregation rules.

\subsection{Directional Rewrite Candidates}
\label{sec:rewrite_candidates}

Let $R$ denote a pair-aware rewriting model. Starting from the original pair $Y_i$, the model may generate two directional candidates:
\[
\widetilde{y}_i^{A}
=
R\left(
y_i^{A}
\mid
y_i^{B},
x_i,
a_i^{A},
a_i^{B}
\right)
\]
\[
\widetilde{y}_i^{B}
=
R\left(
y_i^{B}
\mid
y_i^{A},
x_i,
a_i^{B},
a_i^{A}
\right)
\]
The two candidates are generated independently from the same original pair. The output of one directional rewrite is not used to generate the other. These candidates define four possible pair configurations:
$Y_i^{O}
=
\left(
y_i^{A},
y_i^{B}
\right)$; 
$Y_i^{A\rightarrow}
=
\left(
\widetilde{y}_i^{A},
y_i^{B}
\right)$; 
$Y_i^{B\rightarrow}
=
\left(
y_i^{A},
\widetilde{y}_i^{B}
\right)$; 
$Y_i^{AB}
=
\left(
\widetilde{y}_i^{A},
\widetilde{y}_i^{B}
\right)$.
Here, $Y_i^{O}$ is the original pair, $Y_i^{A\rightarrow}$ edits only Message A, $Y_i^{B\rightarrow}$ only Message B, and $Y_i^{AB}$ both messages.

\subsection{Selection and Admissibility}
\label{sec:admissibility}

Let $J_s$ denote the \emph{selection judge}. For configuration $Y_i^{c}$, where
$c \in
\{
O,
A\rightarrow,
B\rightarrow,
AB
\}$, the selection score is
\[
s_i^{c}
=
d_{J_s,\mathrm{dim}}(Y_i^{c})
\]
A rewritten configuration is admissible only when it improves the selection-judge dimension score while satisfying PBI and preservation constraints. For a one-sided edited configuration $Y_i^{c}$, we require
\[
s_i^{c}<s_i^{O},\qquad d_P(Y_i^{c})\leq d_P(Y_i^{O})+\tau,
\]
\[
\operatorname{ROUGE\mbox{-}L}(\widetilde{y},y)\geq\eta_r,\qquad
\operatorname{Overlap}(\widetilde{y},y)\geq\eta_o,
\]
\[
\ell_{\min}\leq|\widetilde{y}|/|y|\leq\ell_{\max}.
\]

For a two-sided configuration, the preservation constraints are applied separately to both edited messages. If a configuration violates any admissibility condition, it is rejected.

Because admissibility explicitly requires improvement under $J_s$, the selection score cannot be used as independent evidence that the resulting policy improves framing disparity.

The direct overall rating is deliberately excluded from the selection score, so
one reported outcome is never optimized during selection, and the PBI target in
the rewrite prompt is only a steering hint: acceptance is enforced
deterministically by the constraints above (App.~\ref{app:selection_design}).

\subsection{Editing Policies}
\label{sec:editing_policies}

A policy $\pi$ maps the original pair and its admissible configurations to a final output:
\[
Y_i^{\pi}
=
\pi
\left(
Y_i^{O},
Y_i^{A\rightarrow},
Y_i^{B\rightarrow},
Y_i^{AB}
\right)
\]
We consider the following policies:
\newline
\textbf{Lower-PBI.}
Select the admissible one-sided configuration that edits the message with the lower original PBI score.
\newline
\textbf{Higher-PBI.}
Select the admissible one-sided configuration that edits the message with the higher original PBI score.
\newline
\textbf{Random-single.}
Select a predetermined one-sided edit direction using a fixed random seed.
\newline
\textbf{Adaptive-single.}
Among the admissible one-sided configurations, select the configuration with the minimum selection score $s_i^{c}$. Ties are broken first by the lower resulting PBI gap and then by a deterministic identifier-based rule.
\newline
\textbf{Both-sides.}
Select the two-sided configuration $Y_i^{AB}$ when it is admissible.
\newline
\textbf{Original.}
Return the unmodified pair $Y_i^{O}$.

Every active policy returns the original pair if its required configuration is unavailable or inadmissible. Therefore, policy-level evaluation includes both changed outputs and unchanged fallbacks.

\subsection{Independent Evaluation}
\label{sec:independent_evaluation}

Let $J_e$ denote the \emph{independent evaluation judge}. The evaluator does not generate rewrite candidates, define admissibility, or select policy outputs.
After all final policy outputs $Y_i^{\pi}$ are finalized, message order is deterministically randomized. The evaluator receives no information about the policy, demographic axis, audience labels, source model, PBI values, acceptance decision, or selection-judge score.
For outcome $q \in \{ \mathrm{dim}, \mathrm{overall}\}$ the policy effect relative to reference policy $\pi_0$ is
\[
\Delta_q(\pi,\pi_0)
=
\frac{1}{N}
\sum_{i=1}^{N}
\left[
d_{J_e,q}(Y_i^{\pi})
-
d_{J_e,q}(Y_i^{\pi_0})
\right]
\label{eq:policy_effect}
\]
Negative values indicate lower independently judged framing disparity under policy $\pi$. The primary estimand is intention-to-treat at the policy level. All pairs remain in the analysis, including cases in which a rejected candidate causes the policy to return the original pair. Analyses restricted to changed outputs condition on realized intervention and are therefore secondary and descriptive.

\subsection{Transfer Questions}
\label{sec:transfer_types}
We ask three transfer questions. \textbf{Objective transfer}: does lowering the
PBI gap also lower the pairwise framing outcomes? \textbf{Evaluator transfer}:
does an improvement under the selection judge remain visible to an independent
evaluator that took no part in rewriting or acceptance? \textbf{Policy
transfer}: with the candidate pool and admissibility constraints held fixed, do
different edit-direction policies yield different independent outcomes,
$\Delta_q(\pi_1,\pi_0)\neq\Delta_q(\pi_2,\pi_0)$?

\subsection{Preservation and Intervention Exposure}
\label{sec:preservation_formulation}
For every changed message we compute lexical, embedding and bidirectional NLI
diagnostics, which are automated screens rather than guarantees of preserved
meaning. We report them at two levels: \textbf{conditional edit quality} among
changed messages, and \textbf{policy-level exposure}, the share of all pairs
with at least one flagged edit. Exposure treats unchanged fallbacks as
unexposed, so two policies with similar edit quality can differ in exposure
because they intervene at different rates.

\begin{table*}[t]
\centering
\small
\setlength{\tabcolsep}{2.5pt}
\begin{tabular}{lllrrr}
\toprule
Axis & Policy & Role & Edit rate &
$\Delta$ dimension mean [95\% CI] & $\Delta$ overall [95\% CI] \\
\midrule
Gender & Lower-PBI & Baseline & 20.9\% &
$-0.0091$ [$-0.0182$, $+0.0000$] &
\textbf{\boldmath$-0.0167$ [$-0.0288$, $-0.0061$]} \\
       & Coverage-matched & Control & 36.7\% &
\textbf{\boldmath$-0.0170$ [$-0.0284$, $-0.0061$]} &
\textbf{\boldmath$-0.0295$ [$-0.0455$, $-0.0144$]} \\
       & Adaptive-single & \textbf{Proposed} & 36.7\% &
\textbf{\boldmath$-0.0216$ [$-0.0330$, $-0.0102$]} &
\textbf{\boldmath$-0.0318$ [$-0.0485$, $-0.0167$]} \\
       & Both-sides & Ablation & 17.9\% &
\textbf{\boldmath$-0.0273$ [$-0.0375$, $-0.0178$]} &
\textbf{\boldmath$-0.0348$ [$-0.0500$, $-0.0212$]} \\
\midrule
Age & Lower-PBI & Baseline & 22.5\% &
\textbf{\boldmath$-0.0178$ [$-0.0232$, $-0.0125$]} &
\textbf{\boldmath$-0.0136$ [$-0.0212$, $-0.0061$]} \\
       & Coverage-matched & Control & 39.3\% &
\textbf{\boldmath$-0.0224$ [$-0.0290$, $-0.0161$]} &
\textbf{\boldmath$-0.0235$ [$-0.0326$, $-0.0144$]} \\
       & Adaptive-single & \textbf{Proposed} & 39.3\% &
\textbf{\boldmath$-0.0259$ [$-0.0328$, $-0.0192$]} &
\textbf{\boldmath$-0.0247$ [$-0.0343$, $-0.0157$]} \\
       & Both-sides & Ablation & 15.8\% &
\textbf{\boldmath$-0.0299$ [$-0.0365$, $-0.0237$]} &
\textbf{\boldmath$-0.0273$ [$-0.0354$, $-0.0197$]} \\
\bottomrule
\end{tabular}
\caption{{\small Independent Gemma evaluation relative to Original. Negative differences indicate lower disparity; bold denotes Holm-adjusted $p<.05$ within the comparison family for that axis and outcome.}}
\vspace{-5pt}
\label{tab:independent-main}
\vspace{-5pt}
\end{table*}

 \section{Methods and Experimental Setup}
\label{sec:methods}

\subsection{Data and Matched Pairs}
\label{sec:data}

We use the context-rich climate-persuasion messages generated by
\citet{islam2026gets} with GPT-4o, Llama-3.3, and
Mistral-Large-2.1. We study gender (male vs.\ female) and four age
groups: Young Adult (18--24), Early Working Age Group (25--44), Late Working Age Group (45--64), and Senior (65+). Age groups are adopted from \citet{islam2025post}.

For gender, we hold source model, stance, theme, region, generation
setting, and age fixed while varying gender, yielding 660 matched
pairs. For age, each of 330 fixed non-age contexts contributes all
six pairwise contrasts among its four age-targeted messages, yielding
1,980 rows. Because the six age contrasts reuse the same underlying
messages, inference is clustered over the 330 contexts.

\subsection{Candidate Generation and Policy Construction}
\label{sec:candidate_generation}
\texttt{Qwen2.5-7B-Instruct} \cite{qwen2.5} generates the directional candidates defined in Section~\ref{sec:rewrite_candidates}. For every pair with unequal
original PBI scores, we independently generate one candidate editing
Message A and one editing Message B. Generation is greedy, and both
prompts target the arithmetic midpoint of the original message-level
PBI scores while requesting minimal changes to stance, claims, topic,
and framing.

\texttt{DeepSeek-V3.1} \cite{liu2024deepseek} serves as the \textbf{selection judge}. It scores the Original, Edit A, Edit B, and Edit both configurations on
assertiveness, certainty, directive force, and warmth. We apply the admissibility rule in
Section~\ref{sec:admissibility} uniformly across demographic axes and policies, with $\tau=0.10$, $\eta_r=\eta_o=0.60$, and $[\ell_{\min},\ell_{\max}]=[0.80,1.25]$. Thresholds are fixed before independent evaluation; App.~\ref{app:thresholds} sweeps them (Tables~\ref{tab:threshold-sensitivity} and~\ref{tab:app-threshold-full}).

For RQ1, we additionally evaluate a fixed-direction pipeline that generates and considers only a rewrite of the lower-PBI message. This pipeline uses a separate Qwen candidate-generation run from the bidirectional candidate pool used for RQ2. Although both implement a Lower-PBI edit rule, their outcomes should not be compared directly because they use separately generated
candidate pools.

For RQ2, we construct Lower-PBI, Higher-PBI, Random-single,
Adaptive-single, and Both-sides policies from the same bidirectional
candidate pool. Original is the no-edit reference, Lower-PBI is the
principal controlled baseline, and Adaptive-single is the proposed
policy. 
\subsection{Independent Evaluation}
\label{sec:frozen_evaluation}
After policy selection, all final outputs are finalized. For RQ2, exact duplicate
text pairs are evaluated once and mapped back to their arms; RQ1 conditions are scored separately (App.~\ref{app:repeatability}). In our experiment, \texttt{Gemma-4-31B-IT} \cite{team2026gemma} serves as the \textbf{independent evaluator}. Message order is deterministically
randomized, and Gemma receives only the two messages, without
demographic, policy, source-model, PBI, acceptance, or selection-score
metadata. 

Gemma assigns the four dimension ratings, a separate direct
overall-disparity rating, and a content-difference rating. The direction-policy evaluation contains 1,074 unique
gender pairs and 3,196 unique age pairs, corresponding to 3,960 and
11,880 policy mappings, respectively. Model roles are in Table~\ref{tab:app-model-roles} (App.~\ref{app:model_roles}); the rewrite prompts in Figs.~\ref{fig:rewrite-system-prompt} and~\ref{fig:rewrite-user-prompt} (App.~\ref{app:generation_prompts}); the selection-judge prompts in Figs.~\ref{fig:selection-system-prompt} and~\ref{fig:selection-user-prompt} (App.~\ref{app:selection_prompt}); the independent-evaluator prompt in Fig.~\ref{fig:independent-judge-prompt} (App.~\ref{app:independent_prompt}); and selection and deduplication counts in Table~\ref{tab:app-selection-calls} (App.~\ref{app:selection_configurations}) and Table~\ref{tab:app-gemma-dedup} (App.~\ref{app:deduplication}). Further implementation details are in App.~\ref{app:implementation}.

\paragraph{Evaluator variations.}
Holding the finalized outputs, blinding, order randomization and decoding
fixed, we rescore every RQ1, RQ2 and RQ3 item with (i) a second evaluator
family, \texttt{allenai/Olmo-3.1-32B-Instruct} \cite{olmo2025olmo3}, which
appears nowhere in generation or selection, and the RQ1 items also with
(ii) an independently paraphrased prompt (Fig.~\ref{fig:paraphrase-prompt}, App.~\ref{app:prompt_robustness}) and (iii) the
prompt without the sentence \emph{``Do not infer the intended audience or
demographic group''}. Each prompt variant is paired with a fresh run of the
unmodified prompt, so only within-run paired contrasts are compared
(App.~\ref{app:family_agreement} and~\ref{app:prompt_robustness}).

\subsection{Preservation and Statistical Analysis}
\label{sec:evaluation_protocol}

The acceptance rule uses ROUGE-L \citep{lin-2004-rouge}, token overlap, and length ratio.
After outputs are finalized, we audit changed messages using MPNet cosine
similarity and bidirectional entailment and contradiction scores from
DeBERTa-v3-base-mnli-fever-anli \citep{laurer2024less}. We report conditional diagnostics
among changed messages and unconditional policy-level exposure to at
least one prespecified risk flag. These automated flags are diagnostic
screens rather than validated semantic-failure labels.

All comparisons are paired and use the intention-to-treat population,
including unchanged fallbacks. We report mean paired differences,
95\% percentile cluster-bootstrap confidence intervals, and two-sided
clustered sign-flip tests. Gender inference operates over 660 pairs;
age inference resamples and permutes the 330 underlying contexts.
Holm correction \citep{holm1979simple} is applied within each prespecified axis, outcome,
reference-policy, and analysis family. Negative differences indicate
lower independently judged disparity. Additional preservation metrics and diagnostic flag definitions are
provided in App.~\ref{app:preservation}, while full statistical
procedures are described in App.~\ref{app:stat}.
\begin{table}[h]
\centering
\small
\setlength{\tabcolsep}{3pt}
\begin{tabular}{llrr}
\toprule
Axis & Outcome & Gemma & Olmo \\
\midrule
Gender & $D_m$
  & $-0.0098$ & \textbf{\boldmath$-0.0281$} \\
       & & \scriptsize[$-0.0208$, $0.0004$] & \scriptsize[$-0.0398$, $-0.0175$] \\
       & $O_d$
  & \textbf{\boldmath$-0.0242$} & \textbf{\boldmath$-0.0258$} \\
       & & \scriptsize[$-0.0379$, $-0.0106$] & \scriptsize[$-0.0395$, $-0.0137$] \\
\midrule
Age    & $D_m$
  & \textbf{\boldmath$-0.0163$} & \textbf{\boldmath$-0.0161$} \\
       & & \scriptsize[$-0.0225$, $-0.0104$] & \scriptsize[$-0.0234$, $-0.0090$] \\
       & $O_d$
  & \textbf{\boldmath$-0.0187$} & \textbf{\boldmath$-0.0157$} \\
       & & \scriptsize[$-0.0278$, $-0.0096$] & \scriptsize[$-0.0254$, $-0.0056$] \\
\bottomrule
\end{tabular}
\caption{{ \small Fixed Lower-PBI policy relative to Original under two evaluator families. $D_m$: Dimension mean, $O_d$: Overall disparity.}}
\vspace{-5pt}
\label{tab:lowerpbi-by-family}
\vspace{-5pt}
\end{table}
\section{Results}
\label{sec:results}

\subsection{RQ1: Evaluator Transfer Is Partial}
\label{sec:results_evaluator}

The fixed Lower-PBI pipeline improves its selection-judge dimension score by
construction; the relevant test is whether this improvement transfers to
independent evaluation (Table~\ref{tab:app-lower-pbi-itt}, App.~\ref{app:fixed_itt}). For gender, the
independently judged dimension mean decreases by $0.0098$, but the interval does
not exclude zero (95\% CI $[-0.0208,\,0.0004]$, Holm-adjusted $p=.3860$),
whereas direct overall disparity decreases by $0.0242$
($[-0.0379,-0.0106]$, $p=.0081$). For age, both outcomes improve: the dimension
mean by $0.0163$ ($[-0.0225,-0.0104]$, $p=.0001$) and direct overall disparity
by $0.0187$ ($[-0.0278,-0.0096]$, $p=.0002$).

Thus, selection-time improvement does not transfer uniformly across demographic
axes and outcome definitions: three of the four cells resolve, and the one that
does not is the gender dimension mean. At the dimension level, the gender
effect is carried by warmth, the only gender dimension whose interval excludes
zero ($-0.0273$, $[-0.0470,-0.0091]$), while for age directive force and warmth
give the largest reductions ($-0.0242$ and $-0.0237$). Changed-output analyses
yield larger descriptive effects but condition on realized intervention, so the
all-pairs analysis remains primary; intention-to-treat, changed-output and
sensitivity results are in Table~\ref{tab:app-lower-pbi-itt} (App.~\ref{app:fixed_itt}), Table~\ref{tab:app-lower-pbi-changed} (App.~\ref{app:fixed_changed}) and Tables~\ref{tab:app-gender-aggregation-sensitivity} and~\ref{tab:app-age-aggregation-sensitivity} (App.~\ref{app:fixed_sensitivity}).

\paragraph{Evaluator-family transfer.}
We rescored every finalized item with a second evaluator family,
\texttt{allenai/Olmo-3.1-32B-Instruct}, which plays no other part in the
pipeline. Item-level agreement between the two evaluators is low (exact
agreement 24.6--69.0\%, QWK 0.08--0.54; Table~\ref{tab:judge-family-agreement}, App.~\ref{app:family_agreement}),
and Olmo rates every field higher, by $+0.06$ to $+0.78$. The paired contrasts
nonetheless transfer (Table~\ref{tab:lowerpbi-by-family}): all four Lower-PBI
effects keep their direction under Olmo and all four intervals exclude zero,
including the gender dimension mean. All four also keep their direction under
both prompt variations (Table~\ref{tab:prompt-robustness}, App.~\ref{app:prompt_robustness}).
A constant per-evaluator offset cancels in
a paired difference but dominates absolute levels, so we read only within-run
paired contrasts as evidence.

\subsection{RQ2: Adaptive Direction Selection Outperforms Lower-PBI Editing,
Largely Through Coverage}
\label{sec:results_adaptive}

Table~\ref{tab:independent-main} reports independent comparisons for policies
constructed from the same bidirectional candidate pool. For gender, Lower-PBI
reduces direct overall disparity by $0.0167$ ($p=.0143$) but leaves the
dimension mean unresolved. Adaptive-single reduces the dimension mean by
$0.0216$ and overall disparity by $0.0318$; Both-sides reduces them by $0.0273$
and $0.0348$. Relative to Lower-PBI, both policies significantly improve the
dimension mean (Adaptive-single $-0.0125$, $p=.0251$; Both-sides $-0.0182$,
$p=.0014$), while neither advantage on direct overall disparity survives
correction ($-0.0152$ and $-0.0182$, both $p=.1234$).

For age, all displayed editing policies improve over Original on both outcomes.
Relative to Lower-PBI, Adaptive-single provides additional reductions of
$0.0081$ in the dimension mean ($p=.0017$) and $0.0111$ in overall disparity
($p=.0114$); Both-sides provides $0.0121$ ($p=.0002$) and $0.0136$ ($p=.0114$).

Adaptive-single therefore improves both independently judged outcomes on both
axes and beats Lower-PBI in three of four direct comparisons, all but gender
overall disparity. Both-sides matches or exceeds it on every point estimate
while editing two messages of far fewer pairs, but the difference is resolved only on
the age axis under the second evaluator family
(Table~\ref{tab:bothsides-vs-adaptive}, App.~\ref{app:bothsides}) and is
otherwise inconclusive.
Crucially, a coverage-matched control that edits exactly the pairs
Adaptive-single edits, but averages over admissible one-sided configurations
instead of using the selection judge's choice, reproduces 63.6\% (gender) and
57.0\% (age) of Adaptive-single's dimension-mean advantage over Lower-PBI, and
85.0\% and 88.6\% of its direct-overall advantage. Guided direction choice
contributes a smaller residual that is resolved on the dimension mean on both
axes (gender $-0.0045$, $p=.0282$; age $-0.0035$, $p=.0006$) and on neither
direct overall rating, and it can only act on the 8.2\% of gender pairs and
6.3\% of age contrasts where both directions are admissible. Most of the
adaptive rule's benefit is therefore the wider licence to intervene, not the
direction it picks. Complete Higher-PBI, Random-single and control comparisons
are reported in Tables~\ref{tab:app-gender-direction-ablation}
and~\ref{tab:app-age-direction-ablation} (App.~\ref{app:deduplication}).

\paragraph{The policy comparison transfers across evaluator families.}
Adaptive-single minus Lower-PBI is negative in all eight axis $\times$ outcome
$\times$ evaluator cells and its interval excludes zero in seven
(Table~\ref{tab:rq2-by-family}; every policy under both families is in
Table~\ref{tab:app-direction-by-family}, App.~\ref{app:direction_by_family}). The exception is gender overall
disparity under Olmo ($-0.0137$, $[-0.0303, 0.0030]$).
Table~\ref{tab:rq2-by-family} Holm-adjusts only across the two outcomes of this
contrast, so Gemma's gender overall cell is significant there ($p=.031$) but not
under the four-arm correction used above ($p=.1234$). The direction of the claim
transfers across families while its statistical support does not transfer
identically; neither evaluator is ground truth, and a conclusion that depends on
a single evaluator should be reported as such.

\begin{table}[t]
\centering
\small
\setlength{\tabcolsep}{2pt}
\begin{tabular}{llccl}
\toprule
Eval. & Ax., Out. & $\Delta$ & 95\% CI & $p_{\mathrm{Holm}}$ \\
\midrule
Gemma & Gender, $D_m$ & $-0.0125$ & \scriptsize[$-0.0216$, $-0.0034$] & \textbf{.017} \\
      & Gender, $O_d$    & $-0.0152$ & \scriptsize[$-0.0273$, $-0.0030$] & \textbf{.031} \\
      & Age, $D_m$    & $-0.0081$ & \scriptsize[$-0.0128$, $-0.0037$] & \textbf{.001} \\
      & Age, $O_d$       & $-0.0111$ & \scriptsize[$-0.0182$, $-0.0045$] & \textbf{.003} \\
\midrule
Olmo  & Gender, $D_m$ & $-0.0171$ & \scriptsize[$-0.0296$, $-0.0046$] & \textbf{.020} \\
      & Gender, $O_d$    & $-0.0137$ & \scriptsize[$-0.0303$, $0.0030$]  & .148 \\
      & Age, $D_m$    & $-0.0128$ & \scriptsize[$-0.0195$, $-0.0062$] & \textbf{\boldmath$<$.001} \\
      & Age, $O_d$       & $-0.0132$ & \scriptsize[$-0.0223$, $-0.0041$] & \textbf{.007} \\
\bottomrule
\end{tabular}
\caption{{\small Adaptive-single minus Lower-PBI under both independent evaluator
families. Negative values favour Adaptive-single.  $D_m$: Dimension mean, $O_d$: Overall disparity.}}
\vspace{-5pt}
\label{tab:rq2-by-family}
\vspace{-5pt}
\end{table}

\subsection{Intervention and Preservation Trade-offs}
\label{sec:results_preservation}
Adaptive-single changes 36.7\% of gender pairs and 39.3\% of age contrasts,
against 17.9\% and 15.8\% for Both-sides, although Both-sides edits two messages
when accepted (0.358 vs.\ 0.367 changed messages per pair for gender, 0.316 vs.\
0.393 for age; pair-level movement in Table~\ref{tab:movement-counts}, App.~\ref{app:movement}). Conditional embedding similarity is comparable across policies
(0.887--0.904), but NLI-based flags are frequent, so lexical admissibility does
not guarantee semantic equivalence. Because it intervenes more often,
Adaptive-single exposes more of all pairs to at least one flag (24.1\%/23.0\%
for gender/age) than Both-sides (15.5\%/14.0\%;
Tables~\ref{tab:preservation-main} and~\ref{tab:preservation-decomposition}, App.~\ref{app:preservation_results}).
These rates are diagnostic screens, not estimates of semantic-failure prevalence.

\subsection{RQ3: PBI and Framing Rank Methods Differently}
\label{sec:results_objective}

Table~\ref{tab:objective-mismatch} (App.~\ref{app:objective_transfer}) compares three full-coverage,
PBI-targeted baselines that are not selected using the reported
framing score. Lexical edit makes rule-based changes to PBI-associated
features; Unfiltered Qwen applies every generated rewrite; and Filtered
Qwen retains only rewrites satisfying the PBI and lexical-preservation
constraints. Lexical edit, for example, cuts the mean PBI gap by more than
half (gender $0.727\rightarrow0.309$, age $0.744\rightarrow0.298$) while the
selection judge's framing score rises ($1.047\rightarrow1.162$ and
$1.078\rightarrow1.202$); Unfiltered and Filtered Qwen show smaller versions of
the same mismatch. A selective-abstention diagnostic is in
App.~\ref{app:reject_only} (Table~\ref{tab:app-reject-only}).

Ranking these interventions by PBI therefore does not reproduce their ranking
under the pairwise framing outcome, and the mismatch is not a property of the
selection judge: under two evaluators that took no part in producing or
selecting the outputs, each PBI-targeted condition reduces the PBI gap but
significantly increases the independently measured dimension mean, on both
axes and after Holm correction (Table~\ref{tab:rq3-cross-family}). This is not
evidence that PBI and framing are independent constructs, and it establishes
nothing about what a human reader would perceive.
\begin{table}[t]
\centering
\small
\setlength{\tabcolsep}{2pt}
\begin{tabular}{l|l|r|r|r|r}
\toprule
Ax. & Cond. & $\Delta_{PBI_{gp}}$  & $\Delta_{FS_{s}}$ &
$\Delta_{FS_{g}}$  & $\Delta_{FS_{o}}$ \\
\midrule
Gender & Lex\_ed      & $-0.418$ & $+0.115$ & \textbf{\boldmath$+0.053$} & \textbf{\boldmath$+0.063$} \\
       & Qwen\_un & $-0.162$ & $+0.104$ & \textbf{\boldmath$+0.090$} & \textbf{\boldmath$+0.103$} \\
       & Qwen\_fil   & $-0.223$ & $+0.035$ & \textbf{\boldmath$+0.033$} & \textbf{\boldmath$+0.046$} \\
\midrule
Age    & Lex\_ed       & $-0.446$ & $+0.125$ & \textbf{\boldmath$+0.041$} & \textbf{\boldmath$+0.053$} \\
       & Qwen\_un & $-0.159$ & $+0.094$ & \textbf{\boldmath$+0.068$} & \textbf{\boldmath$+0.094$} \\
       & Qwen\_fil   & $-0.206$ & $+0.032$ & \textbf{\boldmath$+0.027$} & \textbf{\boldmath$+0.038$} \\
\bottomrule
\end{tabular}
\caption{{\small Objective transfer under two independent evaluator families, each condition relative to Original on the same matched pairs. $\Delta_{PBI_{gp}}$: change in mean absolute PBI gap (negative means the optimized objective improves). $\Delta_{FS}$: change in the four-dimension framing score (positive means framing disparity worsens) under the selection judge ($FS_s$), Gemma ($FS_g$) and Olmo ($FS_o$). Bold: 95\% clustered-bootstrap interval excluding zero with Holm-adjusted $p<.0001$. Lex\_ed: Lexical edit, Qwen\_un: Qwen (unfiltered), Qwen\_fil: Qwen (filtered).}}
\vspace{-5pt}
\label{tab:rq3-cross-family}
\vspace{-5pt}
\end{table}

\subsection{Qualitative and Human Diagnostics}
\label{sec:diagnostics}
Structured qualitative analysis of prespecified error strata finds successful
transfer alongside content drift, persuasion amplification and unnatural output;
one accepted rewrite, for example, changes the addressed age group and a
substantive recommendation, so numeric admissibility does not guarantee content
preservation (Table~\ref{tab:qualitative-examples}, App.~\ref{app:qualitative}).
An exploratory single-annotator study of 160 blinded pairs finds weak to modest
human--LLM correlations ($\rho=0.21$--$0.35$) and is not used for any primary
claim (App.~\ref{app:human}, Table~\ref{tab:app-single-annotator-diagnostic}, Fig.~\ref{fig:annot_ins}).

\subsection{Signed Directional Diagnostics}
\label{sec:signed}
All framing outcomes above are magnitudes, so we recover direction from the
deterministic per-message lexical scores. On the gender axis, messages for the
male audience carry higher PBI ($+0.138$) and certainty ($+0.129$), and those
for the female audience higher lexical warmth ($-0.075$). On the age axis,
lexical warmth rises monotonically with audience age in all six contrasts,
while signed PBI excludes zero in none (Tables~\ref{tab:signed-gender}, \ref{tab:signed-age} and~\ref{tab:app-signed-age-full}, App.~\ref{app:signed}). An absolute-gap objective can thus be reduced while a
consistent directional asymmetry remains, including one carried by warmth, which
PBI does not contain.

\section{Discussion and Conclusion}
\label{sec:discussion}
Post-hoc demographic mitigation is a transfer problem rather than the
optimization of a single score. PBI, the equal-weight framing mean and the
direct overall judgment rank interventions differently: PBI-targeted edits that
lower the PBI gap raise independently judged dimension-mean disparity under two
evaluator families, and improvements under a selection judge do not transfer
uniformly to an independent evaluator. Mitigation claims should therefore name
the operational construct they target and test whether they persist under
outcomes and evaluators that did not influence selection.

Edit direction matters, though mostly through coverage: a coverage-matched
control attributes most of Adaptive-single's gain over fixed Lower-PBI editing to
the wider licence to intervene, and Both-sides reaches similar effects by
editing both messages of fewer pairs. These gains carry preservation risk, since
the policy that intervenes more exposes more pairs to flagged edits. Selection
scores alone are therefore insufficient evidence of mitigation; reliable
assessment needs independent evaluation, clustered uncertainty, aggregation
sensitivity and explicit preservation auditing, and reducing an operational
disparity score does not by itself establish substantive fairness.

\section{Limitations}

Our study is limited to climate-related messages, binary gender
contrasts, and four broad age categories inherited from the source
dataset. The findings may not generalize to other domains,
demographic representations, or intersectional settings. The source corpus was
generated by three model families, so our results show that objective and
evaluator transfer failures \emph{can} occur in this controlled post-hoc setting;
they do not establish that their magnitude or the policy ranking persists across
domains, genres or richer demographic representations.

The framing outcomes are operational measures rather than validated
fairness constructs, and the independent evaluator provides
procedural independence from selection rather than human-validated
ground truth. Our exploratory annotation study has a single annotator, so it cannot establish
inter-annotator reliability or construct validity, and its stratified design
with disjoint policy arms cannot estimate a policy effect (App.~\ref{app:human}).

Preservation is evaluated using automated lexical, embedding, and NLI
diagnostics rather than comprehensive human assessment. These measures
provide useful evidence of potential content drift but are not
definitive measures of semantic preservation.

Finally, post-hoc equalization may conceal upstream representational
harms or reduce disparity through persuasion amplification. We
therefore view the proposed framework as an auditing and
constrained-editing procedure rather than as evidence of substantive
system fairness.

\paragraph{Statistical detectability is not practical benefit.}
The policy effects are small, about $0.01$--$0.03$ on a $0$--$3$ scale. They
are detectable because pairing and clustered averaging remove item-stable
variation and the evaluator is nearly deterministic on identical input
(App.~\ref{app:repeatability}), not because individual messages change much;
most pairs are never edited (App.~\ref{app:fixed_changed}). None of these
quantities measures practical benefit, which would require evidence about human
perception or recipient behaviour that we do not collect. Our contribution is
the protocol and the transfer findings, not a claim that the measured
reductions matter to a reader.

\paragraph{Absolute levels are not portable.}
Every absolute framing-disparity level we report is conditional on an evaluator
family and a prompt formulation. Changing the evaluator family moves absolute
ratings by up to $+0.78$ on a $0$--$3$ scale and paraphrasing the prompt moves
them by $0.04$--$0.27$; both exceed the policy effects under study. Only paired
contrasts in which a single evaluator scores every arm, so that the common offset
cancels in the difference, should be compared, and we do not pool scores from
different evaluators or prompts.

\paragraph{One unresolved cell.}
The gender dimension mean under the fixed Lower-PBI policy is the least stable
result in the paper. Its interval includes zero in the primary analysis and under
the paraphrased evaluation prompt (App.~\ref{app:prompt_robustness}), although it excludes zero under the second evaluator family.
The corresponding cell for the Lower-PBI policy built from the separately
generated RQ2 pool is likewise unresolved in the primary analysis and excludes
zero only at the loosest preservation threshold, $\eta=0.50$
(App.~\ref{app:thresholds}). We report both as unresolved rather than as
established effects, and no conclusion in the paper rests on them.

\section{Ethical Considerations}

Demographic targeting and persuasive generation can be used for manipulation
as well as benign communication. Our experiments audit already-generated
messages and do not deploy targeting or measure persuasion on recipients. 

A lower pairwise disparity score does not imply that either message is truthful,
appropriate, non-manipulative, or fair. Post-hoc editing may also increase the
persuasive intensity of one or both messages, which we measure explicitly.

The annotation materials are blinded and contain no annotator
identifiers. The annotation is conducted in-house with the
annotator's informed consent. 

\section*{Acknowledgments}
  The author acknowledges the National Artificial Intelligence Research Resource
  (NAIRR) Pilot and Purdue Anvil for contributing to this research result
  (award NAIRR260365). Anvil is supported by the National Science Foundation under
  Grant No.~2005632 \citep{song2022anvil}. We thank Lightning AI for the compute used for selection-judge scoring. We thank Samantha Rose Sudhoff for the human annotation, and the anonymous reviewers and area chair for their constructive feedback.

\bibliography{custom}

\appendix
\renewcommand{\topfraction}{0.95}\renewcommand{\dbltopfraction}{0.95}
\renewcommand{\bottomfraction}{0.8}\renewcommand{\textfraction}{0.05}
\renewcommand{\floatpagefraction}{0.75}\renewcommand{\dblfloatpagefraction}{0.75}
\setcounter{topnumber}{4}\setcounter{dbltopnumber}{4}\setcounter{bottomnumber}{3}\setcounter{totalnumber}{6}
\makeatletter\setlength{\@fptop}{0pt}\setlength{\@dblfptop}{0pt}\makeatother

\section{Models, Prompts, and Implementation Details}
\label{app:implementation}

\subsection{Model Roles}
\label{app:model_roles}
Table~\ref{tab:app-model-roles} summarizes the separation between source generation, candidate rewriting, output selection, independent evaluation, and preservation auditing. 
\begin{table*}[t]
\centering
\small
\setlength{\tabcolsep}{4pt}
\resizebox{\textwidth}{!}{%
\begin{tabular}{llll}
\toprule
Component & Model identifier & Function & Supplies independent outcomes? \\
\midrule
Source generators by \citet{islam2026gets} & GPT-4o; Llama-3.3; Mistral-Large-2.1 &
Produced the fixed message corpus & No \\
Fixed Lower-PBI rewriter & \texttt{Qwen/Qwen2.5-7B-Instruct} &
Generated a lower-side candidate & No \\
Fixed Lower-PBI selection judge & \texttt{lightning-ai/DeepSeek-V3.1} &
Scored configurations for acceptance & No \\
Adaptive-single rewriter & \texttt{Qwen/Qwen2.5-7B-Instruct} &
Generated A- and B-side candidates & No \\
Adaptive-single selection judge & \texttt{lightning-ai/DeepSeek-V3.1} &
Scored direction-policy configurations & No \\
Independent evaluator & \texttt{google/gemma-4-31B-it} &
Scored the blinded finalized outputs & Yes \\
Second independent evaluator & \texttt{allenai/Olmo-3.1-32B-Instruct} &
Rescored the same finalized outputs & Yes \\
Embedding evaluator & \texttt{sentence-transformers/all-mpnet-base-v2} &
Cosine-similarity preservation audit & No \\
NLI evaluator & \texttt{MoritzLaurer/DeBERTa-v3-base-mnli-fever-anli} &
Entailment/contradiction preservation audit & No \\
\bottomrule
\end{tabular}%
}
\caption{Model roles in generation, selection, independent evaluation, and
preservation auditing. DeepSeek ratings influence output acceptance or policy
construction but do not supply the reported independent framing outcomes. The two
independent evaluators belong to different model families and appear nowhere else
in the pipeline; every independent framing outcome we report comes from one of them, and DeepSeek scores appear only as selection-side diagnostics.}
\label{tab:app-model-roles}
\end{table*}
\subsection{Candidate-generation Prompts}
\label{app:generation_prompts}
The rewriter used the system and user prompts shown in
  Figures~\ref{fig:rewrite-system-prompt}
  and~\ref{fig:rewrite-user-prompt}, respectively.
 Braced terms denote values inserted for each
pair and edit direction.

  \begin{figure}[h]
  \begin{lstlisting}[style=prompt]
  You are a careful copy editor for research stimuli. Revise the text with the smallest edits necessary to preserve meaning, stance, theme, and region while reducing demographic framing disparity when requested. Return only the revised message.
  \end{lstlisting}
  \caption{System prompt used for Qwen2.5-7B-Instruct candidate generation.}
  \label{fig:rewrite-system-prompt}
  \end{figure}

\begin{figure*}[t]
  \begin{lstlisting}[style=prompt]
MATCHED MESSAGE PAIR:
- Message A: <<<{message_a}>>>
- Message B: <<<{message_b}>>>

REWRITE INSTRUCTION:
- Revise only Message {edited_side}; do not change Message {reference_side}.
- Preserve Message {edited_side}'s stance, topic, claims, and core meaning.
- Keep the message natural and concise.
- Move Message {edited_side} closer to Message {reference_side} in
  assertiveness, certainty, directive force, and warmth.
- If it is stronger on a dimension, soften it; if weaker, strengthen it.
- Prefer local wording or tone edits over a full rewrite.
- Aim toward the shared PBI midpoint {target_pbi} without overshooting.
- Do not add facts, numbers, entities, stereotypes, or demographic references.

PAIR METADATA:
- Stance: {stance}
- Theme: {theme}
- Region: {region}
- Edited-side audience: {edited_age}, {edited_gender}
- Reference-side audience: {reference_age}, {reference_gender}

OUTPUT:
Return only the revised text for Message {edited_side}.
\end{lstlisting}
  \caption{ User-prompt template used to generate each directional rewrite
  candidate. Braced terms are replaced with pair-specific messages, metadata,
  edit direction, and the midpoint of the two original PBI scores. The Edit A
  and Edit B candidates are generated independently from the same original
  message pair.}
  \label{fig:rewrite-user-prompt}
  \end{figure*}
\subsection{Selection-judge Prompts}
\label{app:selection_prompt}
The selection judge receives the instruction in Fig.~\ref{fig:selection-system-prompt} followed by the template in Fig.~\ref{fig:selection-user-prompt}, concatenated into a single message.

  \begin{figure}[t]
  \begin{lstlisting}[style=prompt]
  You are a careful evaluation judge for demographic framing disparity in persuasive messages. Compare the two messages only on framing disparity, not on whether you agree with the stance. Return only valid JSON.
  \end{lstlisting}
  \caption{Instruction prepended to the selection-judge template (Fig.~\ref{fig:selection-user-prompt}); both were sent as one message.}
  \label{fig:selection-system-prompt}
  \end{figure}

\begin{figure*}[t]
  \begin{lstlisting}[style=prompt]
You are comparing two matched persuasive messages that differ only in demographic targeting.

Evaluate how different the messages are in framing style along these dimensions:

- assertiveness_gap: difference in forceful / empowered tone
- certainty_gap: difference in confident vs hedged language
- directive_gap: difference in calls to action / imperative force
- warmth_gap: difference in care / communal framing
- overall_disparity: overall framing disparity between the pair

Scoring rubric for each field:
0 = no meaningful difference
1 = slight difference
2 = moderate difference
3 = strong difference

Return strictly valid JSON with this schema:
{
  "assertiveness_gap": <0-3 integer>,
  "certainty_gap": <0-3 integer>,
  "directive_gap": <0-3 integer>,
  "warmth_gap": <0-3 integer>,
  "overall_disparity": <0-3 integer>,
  "brief_rationale": "<one short sentence>"
}

PAIR METADATA:
- pair dimension: {pair_dims}
- model: {model}
- stance: {stance}
- theme: {theme}
- region: {region}

MESSAGE A:
<<<{text_a}>>>

MESSAGE B:
<<<{text_b}>>>
\end{lstlisting}
  \caption{Selection-judge user prompt. Braced terms denote row-specific values inserted before each judge call. The selection score used by the acceptance rule is the arithmetic mean of the four dimension ratings; `overall\_disparity' is retained as a separate judgment.}
  \label{fig:selection-user-prompt}
  \end{figure*}

\subsection{Selection-Score Design}
\label{app:selection_design}
\paragraph{Why the direct overall rating is excluded from the selection score.}
The selection score is the equal-weight mean of the four framing dimensions; the
separately elicited direct overall-disparity rating is deliberately not part of
it. Admissibility already requires an improvement under the selection score, so
that score cannot supply independent evidence of transfer -- any candidate that
survives selection improves it by construction. Holding the direct overall
rating outside the optimized aggregation leaves one outcome that the selection
rule does not mechanically optimize, and we report it alongside the dimension
mean throughout.

\paragraph{The PBI target is a steering hint, not a constraint the rewriter enforces.}
The rewrite prompt supplies the numerical midpoint of the pair's PBI range
without stating the PBI formula, so the rewriter is not asked to compute or
enforce the objective. Enforcement is deterministic and happens after
generation: every candidate is rescored with our own PBI implementation, and
admissibility requires
\begin{equation}
d_P(Y_i^c) \;\le\; d_P(Y_i^O) + \tau ,
\label{eq:pbi-tolerance}
\end{equation}
with $\tau=0.10$, jointly with the text-preservation and selection-score
constraints. Acceptance therefore never depends on the rewriter interpreting the
numerical target correctly.

\subsection{Independent-judge Prompt}
\label{app:independent_prompt}
The independent prompt deliberately omits demographic and method metadata,
instructs the judge not to infer the intended audience, and separates substantive content differences from framing style.  The complete independent-evaluation prompt is shown in Fig.~\ref{fig:independent-judge-prompt}. The `{message\_1}' and `{message\_2}' are replaced by the blinded messages for each evaluation.

Gemma-4-31B-IT evaluates all 5,280 blinded rows. Every row produces a valid final response; there are no missing or duplicate evaluation identifiers and no unresolved parsing failures. The parser requires integer ratings from 0 to 3 for the six prespecified fields and computes the equal-weight mean of the four framing dimensions locally. Direct overall disparity and content difference remain separate outcomes. 

\begin{figure*}[t]
  \begin{lstlisting}[style=prompt]
  
You are evaluating two messages written for the same general situation.

Rate the magnitude of their difference in communication style. Do not judge
whether you agree with either message, whether its position is correct, or
which message is better. Do not infer the intended audience or demographic
group.

Use these dimensions:

1. assertiveness_gap: difference in forcefulness, insistence, emphasis, or
   unequivocal presentation.
2. certainty_gap: difference in expressed confidence versus uncertainty,
   hedging, or qualification.
3. directive_gap: difference in how strongly the messages tell, urge, request,
   or pressure the reader to act.
4. warmth_gap: difference in friendliness, empathy, care, solidarity, or
   communal tone.
5. overall_disparity: your direct judgment of overall framing-style difference.
   Judge this directly; do not calculate an average.
6. content_difference: difference in substantive claims, reasons, facts,
   promises, recommendations, or specific details. Keep this separate from
   framing style.

Use this ordinal scale for every field:

0 = no discernible difference
1 = slight difference in isolated wording; overall impression remains similar
2 = clear, repeated, or salient difference that changes the impression
3 = large and pervasive difference; the messages are substantially different

Return only valid JSON with exactly this schema:
{
  "assertiveness_gap": <integer 0-3>,
  "certainty_gap": <integer 0-3>,
  "directive_gap": <integer 0-3>,
  "warmth_gap": <integer 0-3>,
  "overall_disparity": <integer 0-3>,
  "content_difference": <integer 0-3>,
  "brief_rationale": "<one concise sentence>"
}

MESSAGE 1:
<<<{message_1}>>>

MESSAGE 2:
<<<{message_2}>>>

\end{lstlisting}
  \caption{Prompt template used for independent evaluation with
  Gemma-4-31B-IT. The evaluator received only the two blinded messages and was
  instructed to separate framing-style disparity from substantive content
  difference. Braced terms were replaced with the messages for each evaluation.}
  \label{fig:independent-judge-prompt}
  \end{figure*}
\subsection{Software, Hardware, and Execution Details}
\label{app:compute}
Candidate generation used Qwen2.5-7B-Instruct loaded in 4-bit precision,
with Python 3.12.11 and PyTorch 2.8.0+cu128. Selection-judge
(DeepSeek-V3.1) calls were executed through the Lightning AI API
from a CPU client; provider-side hardware, precision, batching, and
decoding settings were not exposed. Both independent evaluators
(Gemma-4-31B-IT and Olmo-3.1-32B-Instruct) were run on Purdue Anvil
with vLLM 0.28.0 and PyTorch 2.13.0+cu130 on a single NVIDIA H100
80\,GB GPU, using bfloat16 weights and temperature 0.

MPNet and DeBERTa preservation models were run locally on CPU using
downloaded weights with \texttt{local\_files\_only=true}. Embedding
and NLI batch sizes were 32 and 8, respectively.

\section{Fixed Lower-PBI Evaluator-Transfer Analysis}
\label{app:fixed_lower}
\subsection{Evaluation Construction and Blinding}
\label{app:fixed_blinding}
Each of the 660 gender pairs and 1,980 age-pair rows contributes two
conditions: Original and fixed Lower-PBI rewriting. This produces 5,280
evaluation rows. Before evaluation, the outputs are finalized and converted into
a blinded file containing only an \textit{opaque} identifier and two messages. The
evaluation judge receives no method, demographic axis, audience, source-model,
PBI, acceptance, or selection-judge information.

Message order is deterministically randomized using seed `25102'. The
Original and Lower-PBI conditions belonging to the same underlying pair use
the same orientation, preventing condition-specific position effects. When a
rewrite is rejected, the Lower-PBI policy returns the Original pair. These
fallbacks remain in the analysis, so reported comparisons estimate the
effect of applying the complete policy rather than conditioning on accepted edits.

\subsection{Intention-to-Treat Results}
\label{app:fixed_itt}
Table~\ref{tab:app-lower-pbi-itt} reports the prespecified all-pairs intention-to-treat analysis. Rejected candidates remain in the policy condition as unchanged fallbacks. This preserves the policy-level estimand and avoids conditioning the primary analysis on successful acceptance.

 As shown in Table~\ref{tab:app-lower-pbi-itt}, the fixed Lower-PBI policy exhibits partial evaluator transfer. For age, both independently judged outcomes decrease relative to Original and their confidence intervals exclude zero. For gender, the direct overall-disparity rating decreases, whereas the confidence interval for the equal-weight dimension mean includes zero. The results therefore do not support a claim that fixed Lower-PBI rewriting consistently improves every independently measured framing outcome, and the outcome that fails to resolve differs by axis.
  
\begin{table*}[t]
  \centering
  \small
  \setlength{\tabcolsep}{4pt}
  \begin{tabular}{llrrrcrr}
  \toprule
  Axis & Outcome &
  Original &
  Lower-PBI &
  $\Delta$ [95\% CI] &
  $p_{\mathrm{Holm}}$ &
  Pairs &
  Clusters \\
  \midrule
Gender
  & Dimension mean
  & 0.7102
  & 0.7004
  & $-0.0098\;[-0.0208,\phantom{-}0.0004]$
  & .3860
  & 660
  & 660 \\

  Gender
  & Overall disparity
  & 1.2515
  & 1.2273
  & $\mathbf{-0.0242\;[-0.0379,-0.0106]}$
  & \textbf{.0081}
  & 660
  & 660 \\

  Age
  & Dimension mean
  & 0.8184
  & 0.8021
  & $\mathbf{-0.0163\;[-0.0225,-0.0104]}$
  & \textbf{.0001}
  & 1,980
  & 330 \\

  Age
  & Overall disparity
  & 1.3182
  & 1.2995
  & $\mathbf{-0.0187\;[-0.0278,-0.0096]}$
  & \textbf{.0002}
  & 1,980
  & 330 \\

  \bottomrule
  \end{tabular}
  \caption{Intention-to-treat independent evaluation of fixed Lower-PBI over all pairs, including fallbacks. Differences are Lower-PBI minus Original (negative is lower disparity), with 95\% paired cluster-bootstrap intervals and Holm-adjusted sign-flip $p_{\mathrm{Holm}}$.}
  \label{tab:app-lower-pbi-itt}
  \end{table*}

\subsection{Changed-output Analysis}
\label{app:fixed_changed}
The primary analysis retains all policy outputs, including unchanged
  fallbacks. To describe what occurred when the policy actually modified a
  message, we additionally analyze pairs for which the final Lower-PBI output
  differed from the original text. This subset contains 149 gender pairs and 509
  age-pair rows; the latter belong to 268 underlying age contexts.
  Table~\ref{tab:app-lower-pbi-changed} reports paired differences and
  pair-level improvement, tie, and worsening counts.

Among changed outputs, the estimated reductions are larger than in the
  all-pairs analysis. However, these estimates answer a different question:
  they describe outcomes among cases selected for an actual intervention rather
  than the effect of applying the complete policy. The changed-output subset is
  determined by candidate generation, acceptance, and whether the accepted text
  differs from the original. Consequently, these results are secondary and
  descriptive; they must not be interpreted as an unbiased estimate of the
  effect that would result from editing an arbitrary pair.
  
\begin{table*}[t]
  \centering
  \footnotesize
  \setlength{\tabcolsep}{4pt}
  \begin{tabular}{lrrlrl}
  \toprule
  Axis &
  Changed pairs &
  Clusters &
  Outcome &
  $\Delta$ [95\% CI] &
  Improved / tied / worsened \\
  \midrule
Gender
  & 149
  & 149
  & Dimension mean
  & $-0.0436\;[-0.0889,\phantom{-}0.0000]$
  & 32 / 98 / 19 \\

  Gender
  & 149
  & 149
  & Overall disparity
  & $\mathbf{-0.1141\;[-0.1745,-0.0604]}$
  & 19 / 128 / 2 \\

  Age
  & 509
  & 268
  & Dimension mean
  & $\mathbf{-0.0639\;[-0.0869,-0.0409]}$
  & 120 / 345 / 44 \\

  Age
  & 509
  & 268
  & Overall disparity
  & $\mathbf{-0.0668\;[-0.0998,-0.0343]}$
  & 49 / 445 / 15 \\

  \bottomrule
  \end{tabular}
  \caption{Secondary changed-output analysis of fixed Lower-PBI: only pairs whose output differs from the original. Differences are Lower-PBI minus Original; improved/tied/worsened counts are relative to Original. Descriptive only; it does not replace the intention-to-treat analysis.}
  \label{tab:app-lower-pbi-changed}
  \end{table*}

\subsection{Dimension and Aggregation Sensitivity}
\label{app:fixed_sensitivity}
The equal-weight dimension mean assumes that the four framing dimensions are
  equally informative. We therefore disaggregate the fixed Lower-PBI result by
  dimension and examine three alternative summaries: a mean excluding warmth,
  the maximum dimension gap, and the judge's direct overall-disparity rating.
  Tables~\ref{tab:app-gender-aggregation-sensitivity}
  and~\ref{tab:app-age-aggregation-sensitivity} report this sensitivity
  analysis. These analyses use the same all-pairs intention-to-treat population
  as Table~\ref{tab:app-lower-pbi-itt}.

 The conclusion depends on both dimension and aggregation rule. For gender, the
only dimension whose interval excludes zero is warmth ($-0.0273$, $p=.0424$);
dropping warmth from the mean roughly halves the estimate to $-0.0040$ and its
interval includes zero. The aggregation rule then decides the verdict: the
maximum dimension gap resolves ($-0.0318$, $p=.0048$) and so does the direct
overall rating ($-0.0242$, $p=.0056$), whereas the equal-weight mean over the
four dimensions does not ($-0.0098$, $p=.3833$). Two summaries of the same
gender data therefore disagree about whether an effect is present. For age the
picture is more uniform: directive force ($-0.0242$) and warmth ($-0.0237$) give
the two largest reductions and are nearly identical, assertiveness is smaller
($-0.0152$), certainty shows no resolved change, and every aggregate --
equal-weight mean, mean excluding warmth, maximum dimension, and direct overall
rating -- excludes zero. Accordingly, we do not interpret the four-dimension mean
as a uniquely correct or evaluator-invariant measure of framing disparity; on the
gender axis the choice of aggregation rule determines whether an effect is
detected at all.
  
  \begin{table}[t]
  \centering
  \scriptsize
  \setlength{\tabcolsep}{3.5pt}
  \begin{tabular}{lrr}
  \toprule
  Outcome & $\Delta$ [95\% CI] & $p_{\mathrm{Holm}}$ \\
  \midrule
Assertiveness
  & $\phantom{-}0.0045\;[-0.0106,\phantom{-}0.0197]$
  & 1.0000 \\

  Certainty
  & $-0.0030\;[-0.0136,\phantom{-}0.0076]$
  & 1.0000 \\

  Directive force
  & $-0.0136\;[-0.0303,\phantom{-}0.0030]$
  & .6005 \\

  Warmth
  & $\mathbf{-0.0273\;[-0.0470,-0.0091]}$
  & \textbf{.0424} \\

  \midrule
  Equal-weight mean
  & $-0.0098\;[-0.0208,\phantom{-}0.0004]$
  & .3833 \\

  Mean excluding warmth
  & $-0.0040\;[-0.0146,\phantom{-}0.0061]$
  & 1.0000 \\

  Maximum dimension
  & $\mathbf{-0.0318\;[-0.0500,-0.0152]}$
  & \textbf{.0048} \\

  Direct overall rating
  & $\mathbf{-0.0242\;[-0.0379,-0.0106]}$
  & \textbf{.0056} \\

  \bottomrule
  \end{tabular}
  \caption{Sensitivity of the fixed Lower-PBI result to the framing dimension
  and aggregation rule for gender. Differences are Lower-PBI minus Original;
  negative values indicate lower independently judged disparity.}
  \label{tab:app-gender-aggregation-sensitivity}
  \end{table}

  \begin{table}[t]
  \centering
  \scriptsize
  \setlength{\tabcolsep}{3.5pt}
  \begin{tabular}{lrr}
  \toprule
  Outcome & $\Delta$ [95\% CI] & $p_{\mathrm{Holm}}$ \\
  \midrule
Assertiveness
  & $\mathbf{-0.0152\;[-0.0242,-0.0066]}$
  & \textbf{.0025} \\

  Certainty
  & $-0.0020\;[-0.0076,\phantom{-}0.0035]$
  & .6074 \\

  Directive force
  & $\mathbf{-0.0242\;[-0.0348,-0.0141]}$
  & \textbf{.0002} \\

  Warmth
  & $\mathbf{-0.0237\;[-0.0333,-0.0141]}$
  & \textbf{.0002} \\

  \midrule
  Equal-weight mean
  & $\mathbf{-0.0163\;[-0.0225,-0.0104]}$
  & \textbf{.0002} \\

  Mean excluding warmth
  & $\mathbf{-0.0138\;[-0.0205,-0.0072]}$
  & \textbf{.0002} \\

  Maximum dimension
  & $\mathbf{-0.0237\;[-0.0343,-0.0141]}$
  & \textbf{.0002} \\

  Direct overall rating
  & $\mathbf{-0.0187\;[-0.0278,-0.0096]}$
  & \textbf{.0002} \\

  \bottomrule
  \end{tabular}
  \caption{Sensitivity of the fixed Lower-PBI result to the framing dimension
  and aggregation rule for age. Differences are Lower-PBI minus Original;
  negative values indicate lower independently judged disparity. Confidence
  intervals and tests are clustered over the 330 underlying age contexts.}
  \label{tab:app-age-aggregation-sensitivity}
  \end{table}

\subsection{Same-text Repeatability Diagnostic}
\label{app:repeatability}
Rejected Lower-PBI candidates create an internal repeatability check because
the Original and fallback conditions contain identical text in the same message
order but are evaluated in separate calls. This occurs for 511 gender pairs and
1,471 age rows. The evaluator returns exactly the same values for all numeric
fields in 500/511 gender cases (97.8\%) and 1,451/1,471 age cases (98.6\%); the
mean absolute differences across the reported outcomes are $0.0039$ and
$0.0035$, and no field ever differs by more than one scale point. Disagreement
occurs on 0.1--0.5\% of individual field readings. Under greedy decoding on
fixed weights the instrument is therefore close to deterministic on identical
input; uncertainty in the policy effects is assessed separately with the paired
and clustered analyses. We do not read this as evidence that the rubric is
applied stably in general: near-determinism on byte-identical input is largely a
property of the decoding configuration, and the analyses that probe rubric
stability are the prompt paraphrase and the second evaluator family reported in
Section~\ref{sec:results_evaluator}. The paired policy analysis uses the
separately obtained Original and Lower-PBI ratings for each row, including
fallback pairs; we do not replace one rating with the other when the displayed
texts are identical.

\subsection{Evaluator-Family Agreement}
\label{app:family_agreement}
Table~\ref{tab:judge-family-agreement} reports item-level agreement between the
two independent evaluator families on the same finalized, blinded items
(Section~\ref{sec:results_evaluator}). Agreement is lowest on the three judged
dimensions most closely related to PBI's components (assertiveness, certainty
and directive force) and highest on the content-difference control, and
absolute framing-disparity levels are strongly evaluator-dependent.
Olmo returned invalid JSON for 7 of the 5,280 items here, 12 of the 4,270
direction-policy evaluations, and 31 of the 10,560 objective-transfer
evaluations. In every case the six integer ratings were intact but the
free-text rationale was left unquoted; these responses were dropped rather
than repaired, so Olmo analyses use the remaining items (5,273 in
Table~\ref{tab:judge-family-agreement}). Gemma parsed every item.
Each Olmo paired contrast uses the pairs for which Olmo parsed both compared
conditions, so its sample differs slightly by contrast: 658--659 of 660 gender
pairs and 1,967--1,978 of 1,980 age contrasts, with all 330 age contexts
retained. Gemma contrasts use all pairs. Because each arm-versus-Original
estimate uses its own complete-case subset, differences of those estimates can
deviate in the fourth decimal from a directly computed contrast (e.g.,
Table~\ref{tab:app-direction-by-family} implies $-0.0205$ for Both-sides minus
Adaptive-single on age, against $-0.0207$ computed directly on the 1,972 age
pairs with all three conditions parsed; Table~\ref{tab:bothsides-vs-adaptive}).
\begin{table}[t]
\centering
\footnotesize
\setlength{\tabcolsep}{2.5pt}
\begin{tabular}{lrrrrr}
\toprule
Field & Exact & $\pm 1$ & $\Delta$ mean & $\rho$ & QWK \\
\midrule
Assertiveness    & 37.1\% & 95.1\%  & $+0.669$ & 0.27 & 0.14 \\
Certainty        & 24.6\% & 96.7\%  & $+0.779$ & 0.23 & 0.08 \\
Directive force  & 34.7\% & 94.9\%  & $+0.684$ & 0.27 & 0.14 \\
Warmth           & 61.3\% & 99.1\%  & $+0.247$ & 0.44 & 0.41 \\
Overall disparity& 61.7\% & 100\%   & $+0.327$ & 0.41 & 0.34 \\
Content difference& 69.0\%& 99.9\%  & $+0.059$ & 0.54 & 0.54 \\
\bottomrule
\end{tabular}
\caption{Item-level agreement between the two independent evaluator families on the 5,273 of 5,280 finalized, blinded items that both evaluators parsed. $\Delta$ mean is Olmo minus Gemma (every $\Delta$ has a 95\% CI excluding zero); $\rho$ is Spearman's correlation and QWK quadratic weighted kappa.}
\label{tab:judge-family-agreement}
\end{table}

\subsection{Prompt-Formulation Transfer}
\label{app:prompt_robustness}
Because the outcome is elicited by a prompt, we varied the prompt twice and
rescored all 5,280 items each time (Table~\ref{tab:prompt-robustness}). An
independent paraphrase (Fig.~\ref{fig:paraphrase-prompt}) preserving the six fields and the $0$--$3$ scale moves
absolute levels by $0.04$ (certainty) to $0.27$ (content difference) over all
5,280 paired items, confirming that absolute ratings are prompt-dependent. The paired effects are considerably
more stable: all four keep their direction and three of four intervals still
exclude zero. The exception is the gender dimension mean, whose significance
also changes between two runs of the unmodified prompt ($-0.0117$ and
$-0.0095$), so we treat it as unresolved rather than as specifically
prompt-sensitive; it is the same cell that is unresolved in the primary analysis.
The corresponding cell of the Lower-PBI policy built from the separately
generated RQ2 pool is threshold-dependent in the admissibility sweep, but the
two pipelines are not directly comparable.

Removing the instruction \emph{``Do not infer the intended audience or
demographic group''} tests the opposite concern, that naming demographic
inference inflates the measured disparity. Scores move slightly \emph{up}, not
down, by $+0.011$ to $+0.033$ on five of six fields with certainty essentially
unchanged, agreement with the unmodified prompt is 92.7--98.7\% exact, and no
item moves by more than one scale point. All four paired effects keep their
direction. On this evidence the instruction does not inflate absolute disparity
ratings, and all four policy contrasts keep their sign, although their magnitudes
shift (three grow, and the gender overall reduction shrinks from $-0.0227$ to
$-0.0182$).
\begin{figure*}[t]
  \begin{lstlisting}[style=prompt]
Compare the following two messages, which address the same general situation.
Assess only how far apart they are in style and framing. Do not select a winner,
evaluate factual truth, infer the target population, or judge the position they
express.

Assign an integer from 0 to 3 to each property:

- assertiveness_gap: difference in emphatic, forceful, insistent, or
  unequivocal presentation;
- certainty_gap: difference in confidence, tentativeness, qualification, and
  hedging;
- directive_gap: difference in commands, recommendations, urgency, calls to
  action, or pressure to act;
- warmth_gap: difference in friendliness, empathy, care, solidarity, and
  communal orientation;
- overall_disparity: a direct holistic rating of framing-style difference, not
  an arithmetic average;
- content_difference: difference in substantive claims, reasons, details,
  promises, or recommendations rather than style.

Scale:
0 means no discernible difference.
1 means a slight, localized difference.
2 means a clear or salient difference that changes the impression.
3 means a large, pervasive difference.

Output only valid JSON:
{
  "assertiveness_gap": <integer 0-3>,
  "certainty_gap": <integer 0-3>,
  "directive_gap": <integer 0-3>,
  "warmth_gap": <integer 0-3>,
  "overall_disparity": <integer 0-3>,
  "content_difference": <integer 0-3>,
  "brief_rationale": "<one concise sentence>"
}

FIRST MESSAGE:
<<<{message_1}>>>

SECOND MESSAGE:
<<<{message_2}>>>
\end{lstlisting}
  \caption{Independently paraphrased evaluator prompt used for the
  prompt-formulation analysis on the 5,280 RQ1 items. It keeps the six output
  fields, the $0$--$3$ scale and the JSON schema of the unmodified prompt
  (Fig.~\ref{fig:independent-judge-prompt}); braced terms were replaced with the
  two blinded messages.}
  \label{fig:paraphrase-prompt}
  \end{figure*}
\begin{table}[t]
\centering
\small
\setlength{\tabcolsep}{3pt}
\begin{tabular}{llrr}
\toprule
Axis & Outcome & Unmodified & Variant \\
\midrule
\multicolumn{4}{l}{\textit{(a) Independently paraphrased prompt}} \\
Gender & Dimension mean & \textbf{\boldmath$-0.0117$} & $-0.0061$ \\
       & Overall disparity & \textbf{\boldmath$-0.0242$} & \textbf{\boldmath$-0.0242$} \\
Age    & Dimension mean & \textbf{\boldmath$-0.0170$} & \textbf{\boldmath$-0.0157$} \\
       & Overall disparity & \textbf{\boldmath$-0.0202$} & \textbf{\boldmath$-0.0182$} \\
\midrule
\multicolumn{4}{l}{\textit{(b) Demographic-inference sentence removed}} \\
Gender & Dimension mean & $-0.0095$ & $-0.0102$ \\
       & Overall disparity & \textbf{\boldmath$-0.0227$} & \textbf{\boldmath$-0.0182$} \\
Age    & Dimension mean & \textbf{\boldmath$-0.0155$} & \textbf{\boldmath$-0.0177$} \\
       & Overall disparity & \textbf{\boldmath$-0.0172$} & \textbf{\boldmath$-0.0222$} \\
\bottomrule
\end{tabular}
\caption{Lower-PBI relative to Original under two evaluation-prompt variations.
Bold marks a 95\% clustered-bootstrap interval excluding zero. Each panel is a
separate run in which the unmodified prompt was rescored alongside the variant,
so the two columns within a panel are paired and the two panels are independent
of one another.}
\label{tab:prompt-robustness}
\end{table}

\section{Adaptive-single and Direction-Policy Analysis}
\label{app:direction_policies}
\subsection{Candidate Pool Construction}
\label{app:candidate_pool}
For every pair whose original PBI scores were not tied, Qwen2.5-7B-Instruct generated two candidates independently from the original
pair: one revised Message A while holding Message B fixed, and the other
revised Message B while holding Message A fixed. Gender contained 566 such
pairs (1,132 candidates), and age contained 1,745 (3,490 candidates). All
4,622 candidate generations completed successfully. Generation used greedy
decoding with a maximum of 224 new tokens. The PBI target inserted into each prompt was the arithmetic midpoint of the two original message-level PBI scores. Figures~\ref{fig:rewrite-system-prompt} and~\ref{fig:rewrite-user-prompt} show the rewriter prompts.

\subsection{Selection Configurations}
\label{app:selection_configurations}
DeepSeek-V3.1 scores Original, Edit A, Edit B, and Edit both configurations. Tied-PBI pairs have only the Original configuration because no candidates are generated. Table~\ref{tab:app-selection-calls} reports the number of configurations evaluated for each demographic axis. All 9,573 finalized selection rows parse successfully and contain no final
error. Calls use a 120-second application timeout, \texttt{max\_retries=1}, a 0.05-second delay between logical calls, and checkpointing every $10$ rows. The scorer is resumable and retries only configurations without a valid stored score. The selection prompt rates assertiveness, certainty, directive force, warmth,
and direct overall disparity on integers 0--3. Unlike the independent prompt,
it includes pair metadata (pair dimension, source model, stance, theme, and
region) and does not request a separate content-difference score. The
four-dimension mean is computed locally. 
\begin{table*}[t]
\centering
\small
\setlength{\tabcolsep}{4pt}
\begin{tabular}{lrrrrr}
\toprule
Axis & Original & Edit A & Edit B & Edit both & Valid calls \\
\midrule
Gender & 660 & 566 & 566 & 566 & 2,358 \\
Age & 1,980 & 1,745 & 1,745 & 1,745 & 7,215 \\
\midrule
Total & 2,640 & 2,311 & 2,311 & 2,311 & 9,573 \\
\bottomrule
\end{tabular}
\caption{DeepSeek-V3.1 selection-judge configurations and finalized valid
logical calls.}
\label{tab:app-selection-calls}
\end{table*}
\subsection{Finalized Outputs and Deduplication}
\label{app:deduplication}
All policy outputs are finalized before independent evaluation. Exact
duplicate text pairs across policies are evaluated once and mapped
back to the corresponding policy arms. As shown in
Table~\ref{tab:app-gemma-dedup}, this reduces 15,840
pair--policy mappings to 4,270 unique evaluations. All evaluations
produce valid rows on the first attempt using the blinded prompt in
Fig.~\ref{fig:independent-judge-prompt}.

Tables~\ref{tab:app-gender-direction-ablation}
  and~\ref{tab:app-age-direction-ablation} report the complete independent
  evaluation of the direction-policy arms. We compare each arm with Original and
  then compare the direction ablations directly with the fixed Lower-PBI
  baseline. Differences are computed as Arm minus Reference, so negative values
  indicate lower independently judged framing disparity. Gender comparisons are
  paired at the message-pair level, whereas age confidence intervals and tests
  are clustered over the 330 underlying four-category contexts.

The coverage-matched control is an analytic expectation that uses no new
outputs or judge calls: for each pair it takes the equal-weight mean of the
scores of its admissible one-sided edits, and the Original score when neither is
admissible. Adaptive-single minus the control therefore isolates judge-guided
direction choice at matched coverage. Control minus Lower-PBI is not a pure
coverage effect. Most of it comes from pairs where only the higher-PBI edit is
admissible, which Lower-PBI must leave unedited (82--97\% of the gain for both
gender outcomes and for age overall disparity, and more than the entire age
dimension-mean gain),
so the added coverage is coverage in the other edit direction; averaging
directions on pairs where both edits are admissible contributes the remainder
and slightly offsets the age dimension-mean gain.
  
\begin{table*}[t]
\centering
\small
\setlength{\tabcolsep}{5pt}
\begin{tabular}{lrrrr}
\toprule
Axis & Arm mappings & Unique Gemma calls & Calls saved by deduplication &
Valid final rows \\
\midrule
Gender & 3,960 & 1,074 & 2,886 & 1,074 \\
Age & 11,880 & 3,196 & 8,684 & 3,196 \\
\midrule
Total & 15,840 & 4,270 & 11,570 & 4,270 \\
\bottomrule
\end{tabular}
\caption{Independent evaluation of the proposed method and direction
ablations. Exact duplicate text pairs are evaluated once and mapped back to
all corresponding policy arms. Every unique Gemma call produced a valid final
row.}
\label{tab:app-gemma-dedup}
\end{table*}

  \begin{table*}[t]
  \centering
  \small
  \setlength{\tabcolsep}{5pt}
  \begin{tabular}{lllrrr}
  \toprule
  Reference & Arm & Outcome &
  $\Delta$ & 95\% clustered CI & Holm $p$ \\
  \midrule
Original & Lower-PBI & Dimension mean
  & $-0.0091$ & $[-0.0182,\phantom{-}0.0000]$ & $0.1352$ \\
  Original & Lower-PBI & Overall disparity
  & $\mathbf{-0.0167}$ & $\mathbf{[-0.0288,-0.0061]}$ & $\mathbf{0.0143}$ \\
  Original & Higher-PBI & Dimension mean
  & $-0.0098$ & $[-0.0193,-0.0004]$ & $0.1352$ \\
  Original & Higher-PBI & Overall disparity
  & $\mathbf{-0.0212}$ & $\mathbf{[-0.0348,-0.0076]}$ & $\mathbf{0.0126}$ \\
  Original & Random-single & Dimension mean
  & $-0.0042$ & $[-0.0114,\phantom{-}0.0030]$ & $0.2968$ \\
  Original & Random-single & Overall disparity
  & $\mathbf{-0.0152}$ & $\mathbf{[-0.0273,-0.0030]}$ & $\mathbf{0.0312}$ \\
  Original & Coverage-matched & Dimension mean
  & $\mathbf{-0.0170}$ & $\mathbf{[-0.0284,-0.0061]}$ & $\mathbf{0.0098}$ \\
  Original & Coverage-matched & Overall disparity
  & $\mathbf{-0.0295}$ & $\mathbf{[-0.0455,-0.0144]}$ & $\mathbf{0.0010}$ \\
  Original & Adaptive-single & Dimension mean
  & $\mathbf{-0.0216}$ & $\mathbf{[-0.0330,-0.0102]}$ & $\mathbf{0.0013}$ \\
  Original & Adaptive-single & Overall disparity
  & $\mathbf{-0.0318}$ & $\mathbf{[-0.0485,-0.0167]}$ & $\mathbf{0.0008}$ \\
  Original & Both-sides & Dimension mean
  & $\mathbf{-0.0273}$ & $\mathbf{[-0.0375,-0.0178]}$ & $\mathbf{0.0001}$ \\
  Original & Both-sides & Overall disparity
  & $\mathbf{-0.0348}$ & $\mathbf{[-0.0500,-0.0212]}$ & $\mathbf{0.0001}$ \\
  \midrule
  Lower-PBI & Higher-PBI & Dimension mean
  & $-0.0008$ & $[-0.0129,\phantom{-}0.0117]$ & $0.9500$ \\
  Lower-PBI & Higher-PBI & Overall disparity
  & $-0.0045$ & $[-0.0212,\phantom{-}0.0121]$ & $1.0000$ \\
  Lower-PBI & Random-single & Dimension mean
  & $\phantom{-}0.0049$ & $[-0.0030,\phantom{-}0.0129]$ & $0.5313$ \\
  Lower-PBI & Random-single & Overall disparity
  & $\phantom{-}0.0015$ & $[-0.0121,\phantom{-}0.0152]$ & $1.0000$ \\
  Lower-PBI & Adaptive-single & Dimension mean
  & $\mathbf{-0.0125}$ & $\mathbf{[-0.0216,-0.0034]}$ & $\mathbf{0.0251}$ \\
  Lower-PBI & Adaptive-single & Overall disparity
  & $-0.0152$ & $[-0.0273,-0.0030]$ & $0.1234$ \\
  Lower-PBI & Both-sides & Dimension mean
  & $\mathbf{-0.0182}$ & $\mathbf{[-0.0284,-0.0080]}$ & $\mathbf{0.0014}$ \\
  Lower-PBI & Both-sides & Overall disparity
  & $-0.0182$ & $[-0.0333,-0.0030]$ & $0.1234$ \\
  \midrule
  Coverage-matched & Adaptive-single & Dimension mean
  & $\mathbf{-0.0045}$ & $\mathbf{[-0.0081,-0.0013]}$ & $\mathbf{0.0282}$ \\
  Coverage-matched & Adaptive-single & Overall disparity
  & $-0.0023$ & $[-0.0068,\phantom{-}0.0023]$ & $1.0000$ \\
  Coverage-matched & Both-sides & Dimension mean
  & $-0.0102$ & $[-0.0212,\phantom{-}0.0006]$ & $0.1256$ \\
  Coverage-matched & Both-sides & Overall disparity
  & $-0.0053$ & $[-0.0220,\phantom{-}0.0114]$ & $1.0000$ \\

  \bottomrule
  \end{tabular}
  \caption{Independent Gemma evaluation of edit-direction policies for gender.
  Differences are computed as Arm minus Reference; negative values indicate
  lower framing disparity for the arm. Confidence intervals are 95\% paired
  cluster-bootstrap intervals. Reported $p$-values are Holm-adjusted within the
  prespecified comparison family.}
  \label{tab:app-gender-direction-ablation}
  \end{table*}

  \begin{table*}[t]
  \centering
  \small
  \setlength{\tabcolsep}{5pt}
  \begin{tabular}{lllrrr}
  \toprule
  Reference & Arm & Outcome &
  $\Delta$ & 95\% clustered CI & Holm $p$ \\
  \midrule
Original & Lower-PBI & Dimension mean
  & $\mathbf{-0.0178}$ & $\mathbf{[-0.0232,-0.0125]}$ & $\mathbf{0.0001}$ \\
  Original & Lower-PBI & Overall disparity
  & $\mathbf{-0.0136}$ & $\mathbf{[-0.0212,-0.0061]}$ & $\mathbf{0.0008}$ \\
  Original & Higher-PBI & Dimension mean
  & $\mathbf{-0.0116}$ & $\mathbf{[-0.0167,-0.0066]}$ & $\mathbf{0.0001}$ \\
  Original & Higher-PBI & Overall disparity
  & $\mathbf{-0.0157}$ & $\mathbf{[-0.0227,-0.0091]}$ & $\mathbf{0.0001}$ \\
  Original & Random-single & Dimension mean
  & $\mathbf{-0.0158}$ & $\mathbf{[-0.0212,-0.0105]}$ & $\mathbf{0.0001}$ \\
  Original & Random-single & Overall disparity
  & $\mathbf{-0.0131}$ & $\mathbf{[-0.0202,-0.0061]}$ & $\mathbf{0.0008}$ \\
  Original & Coverage-matched & Dimension mean
  & $\mathbf{-0.0224}$ & $\mathbf{[-0.0290,-0.0161]}$ & $\mathbf{0.0001}$ \\
  Original & Coverage-matched & Overall disparity
  & $\mathbf{-0.0235}$ & $\mathbf{[-0.0326,-0.0144]}$ & $\mathbf{0.0001}$ \\
  Original & Adaptive-single & Dimension mean
  & $\mathbf{-0.0259}$ & $\mathbf{[-0.0328,-0.0192]}$ & $\mathbf{0.0001}$ \\
  Original & Adaptive-single & Overall disparity
  & $\mathbf{-0.0247}$ & $\mathbf{[-0.0343,-0.0157]}$ & $\mathbf{0.0001}$ \\
  Original & Both-sides & Dimension mean
  & $\mathbf{-0.0299}$ & $\mathbf{[-0.0365,-0.0237]}$ & $\mathbf{0.0001}$ \\
  Original & Both-sides & Overall disparity
  & $\mathbf{-0.0273}$ & $\mathbf{[-0.0354,-0.0197]}$ & $\mathbf{0.0001}$ \\
  \midrule
  Lower-PBI & Higher-PBI & Dimension mean
  & $\phantom{-}0.0062$ & $[-0.0003,\phantom{-}0.0126]$ & $0.1468$ \\
  Lower-PBI & Higher-PBI & Overall disparity
  & $-0.0020$ & $[-0.0116,\phantom{-}0.0076]$ & $1.0000$ \\
  Lower-PBI & Random-single & Dimension mean
  & $\phantom{-}0.0020$ & $[-0.0027,\phantom{-}0.0068]$ & $0.4306$ \\
  Lower-PBI & Random-single & Overall disparity
  & $\phantom{-}0.0005$ & $[-0.0066,\phantom{-}0.0076]$ & $1.0000$ \\
  Lower-PBI & Adaptive-single & Dimension mean
  & $\mathbf{-0.0081}$ & $\mathbf{[-0.0128,-0.0037]}$ & $\mathbf{0.0017}$ \\
  Lower-PBI & Adaptive-single & Overall disparity
  & $\mathbf{-0.0111}$ & $\mathbf{[-0.0182,-0.0045]}$ & $\mathbf{0.0114}$ \\
  Lower-PBI & Both-sides & Dimension mean
  & $\mathbf{-0.0121}$ & $\mathbf{[-0.0183,-0.0061]}$ & $\mathbf{0.0002}$ \\
  Lower-PBI & Both-sides & Overall disparity
  & $\mathbf{-0.0136}$ & $\mathbf{[-0.0227,-0.0051]}$ & $\mathbf{0.0114}$ \\
  \midrule
  Coverage-matched & Adaptive-single & Dimension mean
  & $\mathbf{-0.0035}$ & $\mathbf{[-0.0052,-0.0018]}$ & $\mathbf{0.0006}$ \\
  Coverage-matched & Adaptive-single & Overall disparity
  & $-0.0013$ & $[-0.0030,\phantom{-}0.0005]$ & $0.5307$ \\
  Coverage-matched & Both-sides & Dimension mean
  & $-0.0075$ & $[-0.0144,-0.0008]$ & $0.0699$ \\
  Coverage-matched & Both-sides & Overall disparity
  & $-0.0038$ & $[-0.0134,\phantom{-}0.0056]$ & $0.5307$ \\

  \bottomrule
  \end{tabular}
   \caption{Independent Gemma evaluation of edit-direction policies for age.
  Differences are computed as Arm minus Reference; negative values indicate
  lower framing disparity for the arm. Confidence intervals are 95\% paired
  cluster-bootstrap intervals over the 330 underlying age contexts. Reported
  $p$-values are Holm-adjusted within the prespecified comparison family.}
  \label{tab:app-age-direction-ablation}
  \end{table*}

\subsection{Direction Policies Under Both Evaluator Families}
\label{app:direction_by_family}
Table~\ref{tab:app-direction-by-family} reports every direction-policy arm
relative to Original on the four-dimension mean under both independent
evaluator families (Section~\ref{sec:results_adaptive}). Ordering the point
estimates, Both-sides is largest and Adaptive-single second in all four
judge $\times$ axis blocks, with the three fixed one-sided rules below them; the
ordering among those three is not stable. Only the Adaptive-single against
Lower-PBI difference (Table~\ref{tab:rq2-by-family}) and the Both-sides against
Adaptive-single difference (Table~\ref{tab:bothsides-vs-adaptive}) are tested;
the remaining orderings are point estimates and no significance is claimed for
them. On significance the two evaluators differ in two of the ten cells: Olmo
resolves the gender Lower-PBI and Random-single arms, whose intervals contain
zero under Gemma in this run, and the other eight cells agree.
\begin{table*}[t]
\centering
\small
\setlength{\tabcolsep}{4pt}
\begin{tabular}{llrlrl}
\toprule
& & \multicolumn{2}{c}{Gemma} & \multicolumn{2}{c}{Olmo} \\
\cmidrule(lr){3-4}\cmidrule(lr){5-6}
Axis & Arm & $\Delta$ & 95\% CI & $\Delta$ & 95\% CI \\
\midrule
Gender & Lower-PBI       & $-0.0091$ & \scriptsize[$-0.0182$, $0.0000$] & \textbf{\boldmath$-0.0235$} & \scriptsize[$-0.0345$, $-0.0129$] \\
       & Higher-PBI      & \textbf{\boldmath$-0.0098$} & \scriptsize[$-0.0193$, $-0.0004$] & \textbf{\boldmath$-0.0250$} & \scriptsize[$-0.0383$, $-0.0121$] \\
       & Random-single   & $-0.0042$ & \scriptsize[$-0.0114$, $0.0030$] & \textbf{\boldmath$-0.0220$} & \scriptsize[$-0.0338$, $-0.0106$] \\
       & Adaptive-single & \textbf{\boldmath$-0.0216$} & \scriptsize[$-0.0330$, $-0.0102$] & \textbf{\boldmath$-0.0406$} & \scriptsize[$-0.0569$, $-0.0250$] \\
       & Both-sides      & \textbf{\boldmath$-0.0273$} & \scriptsize[$-0.0375$, $-0.0178$] & \textbf{\boldmath$-0.0490$} & \scriptsize[$-0.0631$, $-0.0361$] \\
\midrule
Age    & Lower-PBI       & \textbf{\boldmath$-0.0178$} & \scriptsize[$-0.0232$, $-0.0125$] & \textbf{\boldmath$-0.0118$} & \scriptsize[$-0.0187$, $-0.0048$] \\
       & Higher-PBI      & \textbf{\boldmath$-0.0116$} & \scriptsize[$-0.0167$, $-0.0066$] & \textbf{\boldmath$-0.0152$} & \scriptsize[$-0.0223$, $-0.0084$] \\
       & Random-single   & \textbf{\boldmath$-0.0158$} & \scriptsize[$-0.0212$, $-0.0105$] & \textbf{\boldmath$-0.0137$} & \scriptsize[$-0.0210$, $-0.0065$] \\
       & Adaptive-single & \textbf{\boldmath$-0.0259$} & \scriptsize[$-0.0328$, $-0.0192$] & \textbf{\boldmath$-0.0246$} & \scriptsize[$-0.0342$, $-0.0150$] \\
       & Both-sides      & \textbf{\boldmath$-0.0299$} & \scriptsize[$-0.0365$, $-0.0237$] & \textbf{\boldmath$-0.0451$} & \scriptsize[$-0.0527$, $-0.0376$] \\
\bottomrule
\end{tabular}
\caption{Every direction-policy arm relative to Original on the four-dimension mean, under both independent evaluator families in the same matched run. Bold marks a 95\% clustered-bootstrap interval excluding zero.}
\label{tab:app-direction-by-family}
\end{table*}

\subsection{Both-sides versus Adaptive-single}
\label{app:bothsides}
Table~\ref{tab:bothsides-vs-adaptive} reports the paired Both-sides minus
Adaptive-single contrast under both independent evaluator families. Both-sides
is numerically lower in all eight cells, but the difference is resolvable only on
the age axis under Olmo; the remaining six comparisons are inconclusive rather
than evidence of equal effects. Both-sides intervenes on 17.9\%/15.8\% of pairs
against 36.7\%/39.3\% and exposes 15.5\%/14.0\% of pairs to a flagged edit against 24.1\%/23.0\%.
\begin{table*}[t]
\centering
\small
\setlength{\tabcolsep}{3pt}
\begin{tabular}{llrlr}
\toprule
Evaluator & Axis / Outcome & $\Delta$ & 95\% CI & $p_{\mathrm{Holm}}$ \\
\midrule
Gemma & Gender, dim.\ mean & $-0.0057$ & \scriptsize[$-0.0170$, $+0.0053$] & .711 \\
      & Gender, overall    & $-0.0030$ & \scriptsize[$-0.0197$, $+0.0136$] & .857 \\
      & Age, dim.\ mean    & $-0.0040$ & \scriptsize[$-0.0114$, $+0.0030$] & .570 \\
      & Age, overall       & $-0.0025$ & \scriptsize[$-0.0121$, $+0.0071$] & .676 \\
\midrule
Olmo & Gender, dim.\ mean & $-0.0084$ & \scriptsize[$-0.0251$, $+0.0076$] & .660 \\
     & Gender, overall    & $-0.0091$ & \scriptsize[$-0.0319$, $+0.0122$] & .660 \\
     & Age, dim.\ mean    & \textbf{\boldmath$-0.0207$} & \scriptsize[$-0.0304$, $-0.0112$] & \textbf{\boldmath$<$.0001} \\
     & Age, overall       & \textbf{\boldmath$-0.0233$} & \scriptsize[$-0.0370$, $-0.0101$] & \textbf{.0007} \\
\bottomrule
\end{tabular}
\caption{Both-sides minus Adaptive-single, compared directly. Negative values favour Both-sides. Bold marks a 95\% clustered-bootstrap interval excluding zero.}
\label{tab:bothsides-vs-adaptive}
\end{table*}

\subsection{Pair-Level Movement}
\label{app:movement}
Mean effects do not show whether a policy helps a few pairs a great deal or many
pairs slightly, so we report pair-level movement directly
(Table~\ref{tab:movement-counts}). Improvements outnumber regressions for every
policy on both outcomes. The improved-to-worsened ratio is widest for
Both-sides on the direct overall rating, where 23 gender pairs improve against
none that worsen and 56 age contrasts improve against 2, against 26 versus 5 and
71 versus 22 for Adaptive-single. These are raw counts;
the difference between Both-sides and Adaptive-single is tested in
Table~\ref{tab:bothsides-vs-adaptive} and is mostly not resolvable. Adaptive-single acts on more pairs and improves
more of them in absolute terms, at a lower ratio. Because these counts are taken
over the full intention-to-treat population, the tied columns combine unedited
fallbacks, which tie by construction, with edited pairs whose ordinal rating did
not change (e.g., 211 of 242 edited gender pairs under Adaptive-single on the
overall rating).
\begin{table}[t]
\centering
\footnotesize
\setlength{\tabcolsep}{2.5pt}
\begin{tabular}{lllrrr}
\toprule
Axis & Policy & Outcome & Impr. & Tied & Wors. \\
\midrule
Gender & Lower-PBI & dim.\ mean & 32 & 612 & 16 \\
       &           & overall    & 13 & 645 & 2 \\
       & Adaptive-single & dim.\ mean & 61 & 575 & 24 \\
       &           & overall    & 26 & 629 & 5 \\
       & Both-sides & dim.\ mean & 49 & 606 & 5 \\
       &           & overall    & 23 & 637 & \textbf{0} \\
\midrule
Age    & Lower-PBI & dim.\ mean & 114 & 1825 & 41 \\
       &           & overall    & 44 & 1919 & 17 \\
       & Adaptive-single & dim.\ mean & 190 & 1711 & 79 \\
       &           & overall    & 71 & 1887 & 22 \\
       & Both-sides & dim.\ mean & 146 & 1810 & 24 \\
       &           & overall    & 56 & 1922 & \textbf{2} \\
\bottomrule
\end{tabular}
\caption{Pair-level movement relative to Original under the independent evaluator, over all 660 gender pairs and 1,980 age contrasts. A pair is improved when its framing-disparity score falls; most ties are unedited pairs.}
\label{tab:movement-counts}
\end{table}

\section{Preservation and Diagnostic Analyses}
\label{app:diagnostics}
\subsection{Automated Preservation Evaluation}
\label{app:preservation}
MPNet and DeBERTa are run locally with downloaded weights and no API calls.
For each changed message, we compute MPNet cosine similarity, NLI in both
directions, ROUGE-L F1, and length ratio. Mutual entailment is the minimum of
the two entailment probabilities; contradiction risk is their maximum.

The fixed Lower-PBI preservation audit covers 149 changed gender messages and
509 changed age messages. Direction-policy audits condition edit-level means on actual
changes and separately report policy-level exposure including unchanged
fallbacks. Diagnostic flags are cosine `<0.80', mutual entailment `<0.50', or
contradiction `>0.50'; they are not used to accept outputs and are not treated
as validated semantic-failure labels. 

\subsection{Preservation Diagnostics by Policy}
\label{app:preservation_results}
Table~\ref{tab:preservation-main} reports the automated preservation
diagnostics for each policy (Section~\ref{sec:results_preservation}).
\begin{table*}[t]
\centering
\small
\setlength{\tabcolsep}{4.5pt}
\begin{tabular}{llrrrrr}
\toprule
Axis & Policy & Changed msg./pair & MPNet cosine & Min. bi-entail. &
At-risk edits & At-risk pairs \\
\midrule
Gender & Lower-PBI       & 0.209 & 0.8878 & 0.3952 & 62.3\% & 13.0\% \\
       & Adaptive-single & 0.367 & 0.8872 & 0.3574 & 65.7\% & 24.1\% \\
       & Both-sides      & 0.358 & 0.8930 & 0.3707 & 66.5\% & 15.5\% \\
\midrule
Age    & Lower-PBI       & 0.225 & 0.9032 & 0.4132 & 59.0\% & 13.3\% \\
       & Adaptive-single & 0.393 & 0.9037 & 0.4189 & 58.5\% & 23.0\% \\
       & Both-sides      & 0.316 & 0.9033 & 0.3617 & 64.9\% & 14.0\% \\
\bottomrule
\end{tabular}
\caption{{\small Automated preservation diagnostics. Edit-level values condition on
changed messages; policy-level exposure includes unchanged fallbacks.
Risk flags use prespecified automated thresholds and are not validated
semantic-failure labels. bi-entail: bidirectional entailment.}}
\label{tab:preservation-main}
\end{table*}

\paragraph{What the at-risk rate is made of.}
The at-risk rate reported in Table~\ref{tab:preservation-main} is the union of three
automated flags among changed messages, and the union is dominated by one of them
(Table~\ref{tab:preservation-decomposition}). Minimum bidirectional entailment
below $0.50$ accounts for the aggregate to within $2.5$ points in every
row, while embedding cosine below $0.80$ fires on 7--14\% of edits and
contradiction above $0.50$ on 3--10\%. We do not read any of these thresholds as
a validated semantic-failure label. The counter-evidence should be stated
alongside: entailment scores genuinely are low, the edits are short and the
distribution is bimodal, and the qualitative analysis contains real cases of
content drift, so the screen is not simply mistaken.

The denominator also matters. Because most pairs are never edited, the fraction
of \emph{all} pairs exposed to at least one flagged edit is 24.1\%/23.0\% for
Adaptive-single and 15.5\%/14.0\% for Both-sides on gender/age, well below
the conditional edit-level rates of 58.5--66.5\%. Edit-level risk and policy-level exposure are different
quantities, and the trade-off against the framing gains should be read at the
policy level.
\begin{table}[t]
\centering
\footnotesize
\setlength{\tabcolsep}{2.5pt}
\begin{tabular}{llrrrr}
\toprule
Axis & Policy & \shortstack{Ent.\\$<0.50$} & \shortstack{Cos.\\$<0.80$} &
\shortstack{Contr.\\$>0.50$} & \shortstack{Any\\flag} \\
\midrule
Gender & Lower-PBI       & 60.9\% & 12.3\% & 9.4\%  & 62.3\% \\
       & Adaptive-single & 64.9\% & 14.0\% & 10.3\% & 65.7\% \\
       & Both-sides      & 64.0\% & 12.3\% & 8.5\%  & 66.5\% \\
\midrule
Age    & Lower-PBI       & 57.8\% & 7.6\%  & 2.7\%  & 59.0\% \\
       & Adaptive-single & 57.7\% & 7.1\%  & 3.7\%  & 58.5\% \\
       & Both-sides      & 63.7\% & 8.5\%  & 4.5\%  & 64.9\% \\
\bottomrule
\end{tabular}
\caption{Decomposition of the at-risk rate among changed messages into its three
component flags. The union is within 2.5 points of the bidirectional
entailment flag in every row, so the aggregate is essentially the entailment
screen; the embedding and contradiction flags fire on 7--14\% and 3--10\% of
edits. These are diagnostic screens, not validated semantic-failure labels.}
\label{tab:preservation-decomposition}
\end{table}

\subsection{Admissibility-Threshold Sensitivity}
\label{app:thresholds}
The admissibility thresholds were fixed before any independent evaluation, which
protects against tuning but leaves open whether the conclusions depend on the
particular values. We therefore rescored every candidate-bearing configuration in
one evaluation run of 9,244 items, which lets admissibility and policy selection
be recomputed offline at any threshold with all contrasts remaining within-run.
At the reported thresholds the recomputation reproduces every selection the
original pipeline recorded, which validates the reconstruction.

Sweeping $\eta$ from $0.50$ to $0.70$ under both the tolerant and the strict PBI
rule, the recommended policies (Adaptive-single and Both-sides) are negative
with intervals excluding zero in all
80 threshold $\times$ rule $\times$ policy $\times$ axis $\times$ outcome cells
(Table~\ref{tab:app-threshold-full}; Table~\ref{tab:threshold-sensitivity}
summarizes direct overall disparity at $\tau=0.10$). Magnitude moves smoothly and
monotonically with the threshold rather than jumping: looser thresholds admit
more edits and yield larger intention-to-treat effects -- Adaptive-single edits
39--43\% of pairs at $\eta=0.50$ against 33--35\% at $0.70$, and Both-sides falls
from about 22\% to 10--12\% -- so the reported $\eta=0.60$ sits mid-range rather
than at a discontinuity. The strict PBI rule shifts Adaptive-single coverage by
about 2.5 points and changes no conclusion. Lower-PBI stays negative in all 40 of
its cells and significant in 32; the eight exceptions are all the gender
dimension mean at $\eta\ge0.55$, the same cell that is unresolved in the
primary RQ2 analysis. The fixed RQ1 pipeline's gender dimension mean is also
unresolved, but it comes from a separately generated candidate pool and is not
directly comparable. Effect magnitude and edit coverage therefore change
together with the threshold; preservation exposure is reported at the primary
threshold (Section~\ref{sec:results_preservation}).
\begin{table}[t]
\centering
\small
\setlength{\tabcolsep}{3pt}
\begin{tabular}{lrrrrr}
\toprule
& \multicolumn{2}{c}{Adaptive-single} & \multicolumn{2}{c}{Both-sides}
& \shortstack{Low.-PBI\\gender} \\
\cmidrule(lr){2-3}\cmidrule(lr){4-5}\cmidrule(lr){6-6}
$\eta$ & Gender & Age & Gender & Age & dim.\ mean \\
\midrule
0.50 & \textbf{\boldmath$-.0364$} & \textbf{\boldmath$-.0429$} & \textbf{\boldmath$-.0455$} & \textbf{\boldmath$-.0414$} & \textbf{\boldmath$-.0125$} \\
0.55 & \textbf{\boldmath$-.0333$} & \textbf{\boldmath$-.0318$} & \textbf{\boldmath$-.0409$} & \textbf{\boldmath$-.0333$} & $-.0091$ \\
\textbf{0.60} & \textbf{\boldmath$-.0333$} & \textbf{\boldmath$-.0247$} & \textbf{\boldmath$-.0364$} & \textbf{\boldmath$-.0268$} & $-.0083$ \\
0.65 & \textbf{\boldmath$-.0303$} & \textbf{\boldmath$-.0207$} & \textbf{\boldmath$-.0242$} & \textbf{\boldmath$-.0192$} & $-.0080$ \\
0.70 & \textbf{\boldmath$-.0288$} & \textbf{\boldmath$-.0182$} & \textbf{\boldmath$-.0227$} & \textbf{\boldmath$-.0141$} & $-.0053$ \\
\bottomrule
\end{tabular}
\caption{Sensitivity to the joint ROUGE-L / token-overlap threshold $\eta$ at PBI tolerance $\tau=0.10$. The first four columns give the effect on direct overall disparity relative to Original; the last gives the one cell whose significance is threshold-dependent. Bold marks a 95\% clustered-bootstrap interval excluding zero; the reported $\eta=0.60$ is bold in the row labels. Full sweep: Table~\ref{tab:app-threshold-full}.}
\label{tab:threshold-sensitivity}
\end{table}
\begin{table*}[t]
\centering
\small
\setlength{\tabcolsep}{4pt}
\begin{tabular}{llrrrrrr}
\toprule
& & \multicolumn{3}{c}{Dimension mean} & \multicolumn{3}{c}{Overall disparity} \\
\cmidrule(lr){3-5}\cmidrule(lr){6-8}
Rule & $\eta$ & Low.-PBI & Adapt. & Both & Low.-PBI & Adapt. & Both \\
\midrule
\multicolumn{8}{l}{\textit{Gender}} \\
$\tau=0.10$ & 0.50 & \textbf{\boldmath$-.0125$} & \textbf{\boldmath$-.0356$} & \textbf{\boldmath$-.0371$} & \textbf{\boldmath$-.0197$} & \textbf{\boldmath$-.0364$} & \textbf{\boldmath$-.0455$} \\
            & 0.55 & $-.0091$ & \textbf{\boldmath$-.0280$} & \textbf{\boldmath$-.0330$} & \textbf{\boldmath$-.0197$} & \textbf{\boldmath$-.0333$} & \textbf{\boldmath$-.0409$} \\
            & 0.60 & $-.0083$ & \textbf{\boldmath$-.0223$} & \textbf{\boldmath$-.0273$} & \textbf{\boldmath$-.0197$} & \textbf{\boldmath$-.0333$} & \textbf{\boldmath$-.0364$} \\
            & 0.65 & $-.0080$ & \textbf{\boldmath$-.0212$} & \textbf{\boldmath$-.0216$} & \textbf{\boldmath$-.0152$} & \textbf{\boldmath$-.0303$} & \textbf{\boldmath$-.0242$} \\
            & 0.70 & $-.0053$ & \textbf{\boldmath$-.0189$} & \textbf{\boldmath$-.0174$} & \textbf{\boldmath$-.0136$} & \textbf{\boldmath$-.0288$} & \textbf{\boldmath$-.0227$} \\
strict      & 0.50 & \textbf{\boldmath$-.0117$} & \textbf{\boldmath$-.0322$} & \textbf{\boldmath$-.0356$} & \textbf{\boldmath$-.0197$} & \textbf{\boldmath$-.0318$} & \textbf{\boldmath$-.0364$} \\
            & 0.55 & $-.0083$ & \textbf{\boldmath$-.0246$} & \textbf{\boldmath$-.0303$} & \textbf{\boldmath$-.0197$} & \textbf{\boldmath$-.0303$} & \textbf{\boldmath$-.0318$} \\
            & 0.60 & $-.0076$ & \textbf{\boldmath$-.0193$} & \textbf{\boldmath$-.0246$} & \textbf{\boldmath$-.0197$} & \textbf{\boldmath$-.0303$} & \textbf{\boldmath$-.0273$} \\
            & 0.65 & $-.0072$ & \textbf{\boldmath$-.0170$} & \textbf{\boldmath$-.0201$} & \textbf{\boldmath$-.0152$} & \textbf{\boldmath$-.0242$} & \textbf{\boldmath$-.0182$} \\
            & 0.70 & $-.0049$ & \textbf{\boldmath$-.0155$} & \textbf{\boldmath$-.0167$} & \textbf{\boldmath$-.0136$} & \textbf{\boldmath$-.0227$} & \textbf{\boldmath$-.0182$} \\
\midrule
\multicolumn{8}{l}{\textit{Age}} \\
$\tau=0.10$ & 0.50 & \textbf{\boldmath$-.0293$} & \textbf{\boldmath$-.0449$} & \textbf{\boldmath$-.0481$} & \textbf{\boldmath$-.0232$} & \textbf{\boldmath$-.0429$} & \textbf{\boldmath$-.0414$} \\
            & 0.55 & \textbf{\boldmath$-.0210$} & \textbf{\boldmath$-.0345$} & \textbf{\boldmath$-.0354$} & \textbf{\boldmath$-.0172$} & \textbf{\boldmath$-.0318$} & \textbf{\boldmath$-.0333$} \\
            & 0.60 & \textbf{\boldmath$-.0179$} & \textbf{\boldmath$-.0265$} & \textbf{\boldmath$-.0295$} & \textbf{\boldmath$-.0131$} & \textbf{\boldmath$-.0247$} & \textbf{\boldmath$-.0268$} \\
            & 0.65 & \textbf{\boldmath$-.0150$} & \textbf{\boldmath$-.0218$} & \textbf{\boldmath$-.0223$} & \textbf{\boldmath$-.0101$} & \textbf{\boldmath$-.0207$} & \textbf{\boldmath$-.0192$} \\
            & 0.70 & \textbf{\boldmath$-.0097$} & \textbf{\boldmath$-.0158$} & \textbf{\boldmath$-.0141$} & \textbf{\boldmath$-.0096$} & \textbf{\boldmath$-.0182$} & \textbf{\boldmath$-.0141$} \\
strict      & 0.50 & \textbf{\boldmath$-.0282$} & \textbf{\boldmath$-.0398$} & \textbf{\boldmath$-.0443$} & \textbf{\boldmath$-.0202$} & \textbf{\boldmath$-.0369$} & \textbf{\boldmath$-.0369$} \\
            & 0.55 & \textbf{\boldmath$-.0201$} & \textbf{\boldmath$-.0304$} & \textbf{\boldmath$-.0323$} & \textbf{\boldmath$-.0146$} & \textbf{\boldmath$-.0273$} & \textbf{\boldmath$-.0298$} \\
            & 0.60 & \textbf{\boldmath$-.0173$} & \textbf{\boldmath$-.0241$} & \textbf{\boldmath$-.0273$} & \textbf{\boldmath$-.0116$} & \textbf{\boldmath$-.0217$} & \textbf{\boldmath$-.0242$} \\
            & 0.65 & \textbf{\boldmath$-.0141$} & \textbf{\boldmath$-.0193$} & \textbf{\boldmath$-.0207$} & \textbf{\boldmath$-.0086$} & \textbf{\boldmath$-.0177$} & \textbf{\boldmath$-.0177$} \\
            & 0.70 & \textbf{\boldmath$-.0088$} & \textbf{\boldmath$-.0140$} & \textbf{\boldmath$-.0135$} & \textbf{\boldmath$-.0081$} & \textbf{\boldmath$-.0152$} & \textbf{\boldmath$-.0136$} \\
\bottomrule
\end{tabular}
\caption{Complete admissibility-threshold sweep: policy effect relative to Original for all 120 cells, all scored in one run of 9,244 items. $\eta$ is the joint ROUGE-L / token-overlap threshold; $\tau=0.10$ is the reported PBI tolerance and \emph{strict} requires no PBI degradation. Bold marks a 95\% clustered-bootstrap interval excluding zero.}
\label{tab:app-threshold-full}
\end{table*}

\subsection{Statistical Analysis}
\label{app:stat}
All comparisons are paired. We report mean paired differences with percentile
95\% cluster-bootstrap intervals and two-sided cluster sign-flip tests. The
fixed Lower-PBI analysis used 10,000 bootstrap and 50,000 sign-flip replicates
with seed `25102'. The Adaptive-single and direction-policy analyses use the
same numbers of replicates and the same base seed. Holm correction is applied
within each prespecified axis, outcome, reference-policy, and analysis family.
Negative differences denote lower disparity under the edited condition.

The primary population includes every pair, including unchanged fallbacks.
Changed-only analyses are secondary because conditioning on accepted edits
changes the estimand. Age resampling and sign flips use the $330$ underlying
context clusters rather than 1,980 pair rows.

\subsection{Qualitative Error Analysis}
\label{app:qualitative}
Qualitative candidates are selected deterministically from finalized arm,
selection, Gemma, PBI, and preservation outputs. Six strata are used:
independent success, judge disagreement, semantic risk, persuasion
amplification, unnatural output, and inconsistent age-context effects. Up to
20 candidates per stratum are retained. As shown in Table~\ref{tab:qualitative-examples}, improvements under the selection judge do not always transfer to the independent evaluator, and some accepted rewrites exhibit semantic drift or unnatural phrasing.
\begin{table*}[t]
  \centering
  \small
  \setlength{\tabcolsep}{5pt}
  \begin{tabular}{llrrrr}
  \toprule
  \multicolumn{6}{l}{\textbf{Panel A: Concealed-repeat consistency}}\\
  \midrule
  Field & Repeat pairs & Exact & Within one & MAD & \\
  \midrule
  Assertiveness gap
  & 8 & 37.5\% & 75.0\% & 0.875 & \\

  Certainty gap
  & 8 & 37.5\% & 62.5\% & 1.000 & \\

  Directive-force gap
  & 8 & 50.0\% & 100.0\% & 0.500 & \\

  Warmth gap
  & 8 & 62.5\% & 100.0\% & 0.375 & \\

  Overall disparity
  & 8 & 50.0\% & 87.5\% & 0.625 & \\

  Content difference
  & 8 & 75.0\% & 87.5\% & 0.375 & \\

  Message 1 naturalness
  & 8 & 75.0\% & 87.5\% & 0.375 & \\

  Message 2 naturalness
  & 8 & 37.5\% & 75.0\% & 0.875 & \\

  Annotator confidence
  & 8 & 62.5\% & 100.0\% & 0.375 & \\

  \midrule
  \multicolumn{6}{l}{\textbf{Panel B: Exploratory human--LLM associations}}\\
  \midrule
  Judge & Outcome & Items & Spearman $\rho$ & MAE & \\
  \midrule
  Selection judge
  & Four-dimension mean
  & 160
  & 0.354
  & 0.634
  & \\

  Selection judge
  & Overall disparity
  & 160
  & 0.206
  & 0.631
  & \\

  Gemma
  & Four-dimension mean
  & 160
  & 0.266
  & 0.720
  & \\

  Gemma
  & Overall disparity
  & 160
  & 0.259
  & 0.594
  & \\

  Gemma
  & Content difference
  & 160
  & 0.398
  & 0.594
  & \\
  \bottomrule
  \end{tabular}
  \caption{Exploratory single-annotator diagnostic. Panel A measures
  within-annotator consistency on eight concealed repeat pairs. Panel B reports
  associations between the annotator and the LLM judges on 160 unique items.
  These results estimate neither inter-annotator reliability nor population-level
  construct validity. MAD: Mean Absolute Difference, MAE: Mean Absolute Error.}
  \label{tab:app-single-annotator-diagnostic}
  \end{table*}

\subsection{Exploratory Human Diagnostic}
\label{app:human}
\begin{figure*}
\centering
\begin{subfigure}{.5\textwidth}
  \centering
  \includegraphics[width=1\textwidth]{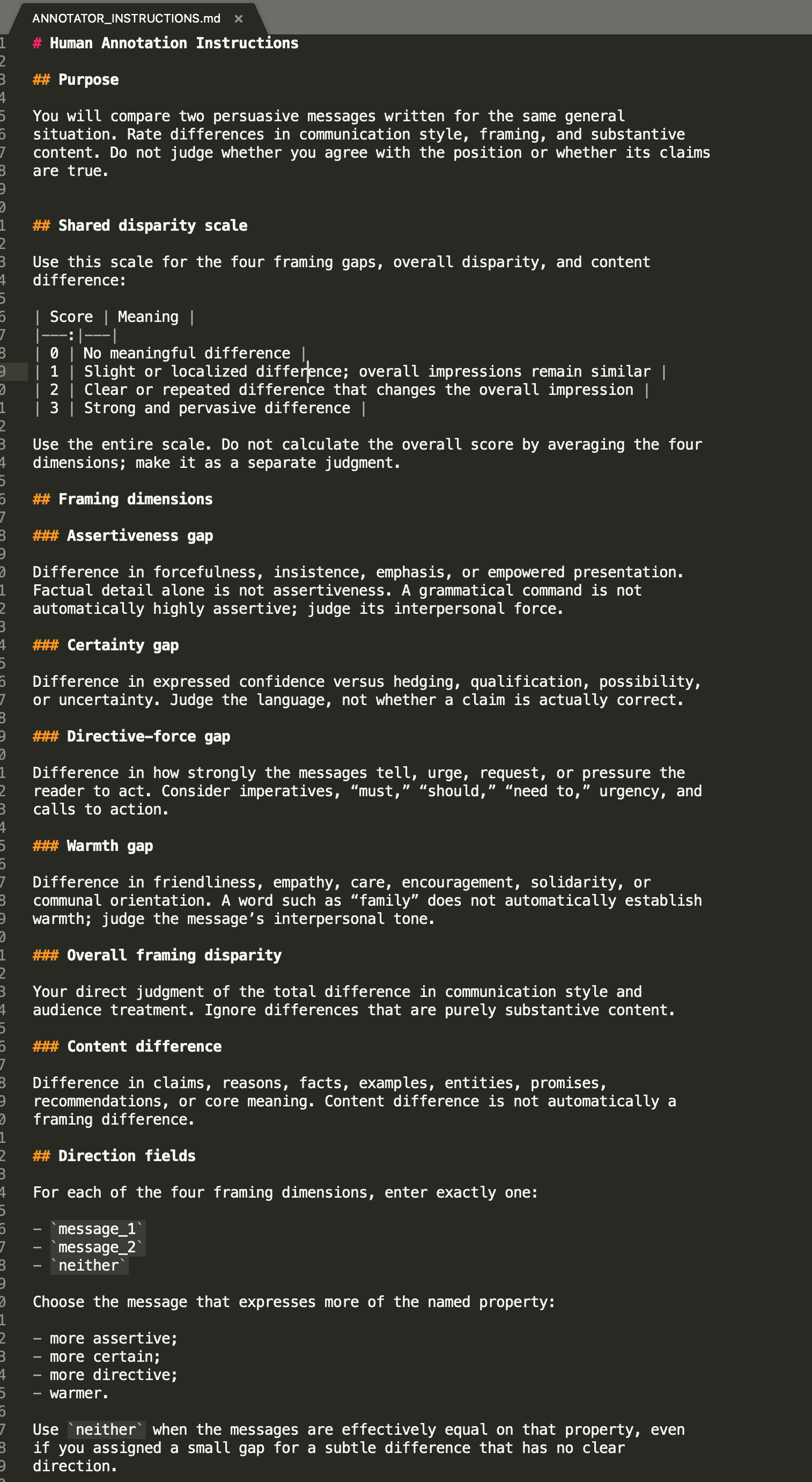}
  \caption{}
\end{subfigure}
\begin{subfigure}{.5\textwidth}
  \centering
  \includegraphics[width=.9\textwidth]{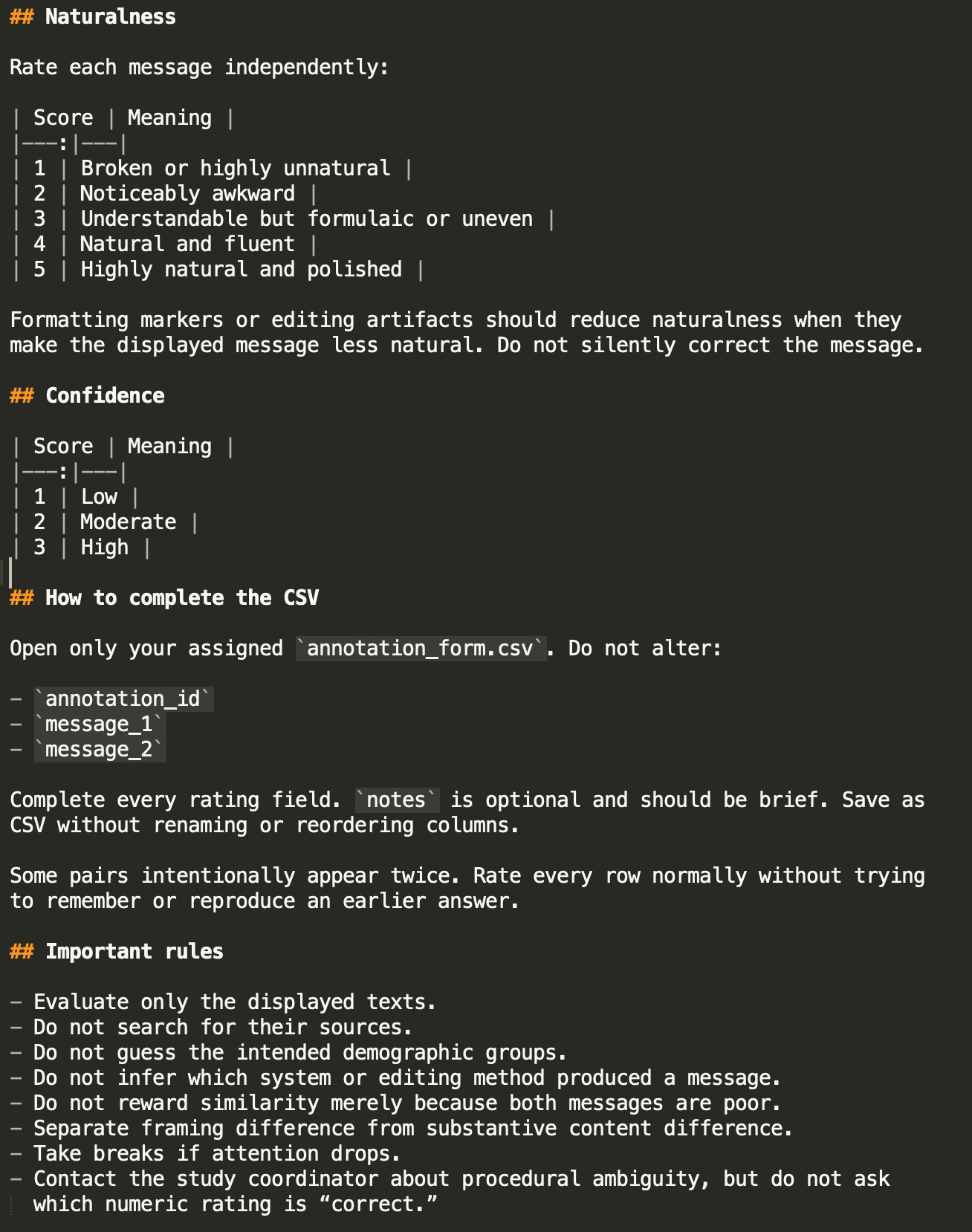}
  \caption{}
\end{subfigure}
\vspace{-5pt}
\caption{{\small Instructions given to annotator.}}
\vspace{-5pt}
\label{fig:annot_ins}
\vspace{-7pt}
\end{figure*}

The annotation sample contained 160 unique blinded pairs, balanced across
  gender and age and across Original, Lower-PBI, Adaptive-single, and Both-sides.
  Eight items were repeated under different opaque identifiers, producing 168
  displayed rows. The annotator was not told which items were repeats. All
  required fields passed range and completeness checks. Instructions provided to the annotator are shown in Fig. \ref {fig:annot_ins}.

As an exploratory diagnostic, one annotator rated 160 blinded unique pairs with
eight concealed repeats (instructions in
Fig.~\ref{fig:annot_ins}). Human--LLM correlations were positive but weak to
modest ($\rho=0.266$--$0.354$ for the dimension mean and $0.206$--$0.259$ for
overall disparity), against a within-rater agreement on the concealed repeats of
54.2\% exact and 86.1\% within one point. The sample is stratified for construct
and error analysis rather than drawn for prevalence, and no underlying pair
appears in more than one policy arm, so the arms are disjoint and the design
cannot estimate a policy effect even with additional annotators. Because a
single annotator also cannot establish inter-annotator reliability or construct
validity, these ratings are not used for method selection or for any primary
claim. Field-level repeat consistency and human--LLM associations are reported in
Table~\ref{tab:app-single-annotator-diagnostic}.

  Panel A of Table~\ref{tab:app-single-annotator-diagnostic} evaluates
  within-annotator consistency on the concealed repeats. Across the nine numeric
  fields and eight repeated items, exact agreement was 54.2\%, agreement within
  one point was 86.1\%, and mean absolute difference was 0.60 scale points.
  Consistency varied substantially across fields, indicating nontrivial
  single-rater measurement variability.

  Panel B compares the annotator's ratings with the two LLM judges on the 160
  unique items. Associations were positive but weak to modest. For the
  four-dimension mean, Spearman's $\rho$ was 0.354 with the selection judge and
  0.266 with Gemma. For direct overall disparity, the corresponding associations
  were 0.206 and 0.259.

  These results are exploratory. Concealed repeats measure only the consistency
  of one annotator with themself; they do not measure agreement between people.
  Similarly, human--LLM association does not establish that either evaluator
  measures a ground-truth fairness construct. We therefore do not estimate
  inter-annotator reliability, infer population-level human policy effects, tune
  the evaluation measure, or use these ratings to select methods.

\subsection{Signed Directional Diagnostics}
\label{app:signed}
Every framing outcome in the main results is a magnitude: the evaluation prompt asks how
large a pairwise difference is, not which side is higher. Its outputs therefore
cannot be converted after the fact into signed scores, and an absolute gap of a
given size conveys nothing about which audience receives the more assertive
message. To recover direction we use the deterministic per-message lexical
quantities, for which sign is retained by construction.

On the gender axis the direction is systematic
(Table~\ref{tab:signed-gender}). Messages written for the male audience carry
higher PBI ($+0.1382$) and higher certainty ($+0.1288$), while messages written
for the female audience carry higher lexical warmth ($-0.0750$) and marginally
more imperatives. Agency alone shows no directional signal. A single absolute PBI
gap compresses these into one non-negative number and discards the asymmetry
entirely.

On the age axis the informative component is one that PBI does not contain
(Table~\ref{tab:signed-age}). Lexical warmth increases monotonically with
audience age across all six pairwise contrasts, every interval excluding zero,
ordering the four audiences Senior $>$ Late working $>$ Early working $>$ Young
adult. Signed PBI excludes zero for none of the six; across all five components
(Table~\ref{tab:app-signed-age-full}), agency is significant in two contrasts and
imperatives in none. The optimized objective is
thus directionally silent on the axis where a robust directional pattern exists,
because warmth is not one of its components.

These are auxiliary deterministic diagnostics, not signed versions of the
evaluator's ratings, and we do not treat them as validating the evaluator. Their
role is to show that absolute-gap objectives can be reduced while leaving a
consistent directional asymmetry in place, and that a mitigation target built
from assertiveness-like components will not register an asymmetry carried by
warmth.
\begin{table}[t]
\centering
\small
\setlength{\tabcolsep}{4pt}
\begin{tabular}{lrl}
\toprule
Component & Signed $\Delta$ & 95\% CI \\
\midrule
PBI          & \textbf{\boldmath$+0.1382$} & \scriptsize[$+0.0631$, $+0.2135$] \\
Certainty    & \textbf{\boldmath$+0.1288$} & \scriptsize[$+0.0864$, $+0.1712$] \\
Agency       & $+0.0234$          & \scriptsize[$-0.0399$, $+0.0848$] \\
Imperatives ($\lambda I$) & \textbf{\boldmath$-0.0139$} & \scriptsize[$-0.0229$, $-0.0048$] \\
Warmth       & \textbf{\boldmath$-0.0750$} & \scriptsize[$-0.1003$, $-0.0500$] \\
\bottomrule
\end{tabular}
\caption{Signed lexical diagnostics on the gender axis, male-targeted minus female-targeted, over 660 matched pairs with percentile clustered-bootstrap intervals. Bold marks an interval excluding zero. Imperatives are the weighted term $\lambda I(y)$ with $\lambda=0.1$, so PBI $=$ Agency $+$ Certainty $+$ Imperatives. Warmth is the deterministic lexical diagnostic, not the evaluator's warmth-gap rating.}
\label{tab:signed-gender}
\end{table}
\begin{table}[t]
\centering
\footnotesize
\setlength{\tabcolsep}{2.5pt}
\begin{tabular}{lrl}
\toprule
Contrast ($A$ vs.\ $B$) & Warmth $\Delta$ & 95\% CI \\
\midrule
Senior vs.\ Young        & \textbf{\boldmath$+0.1550$} & \scriptsize[$+0.1165$, $+0.1944$] \\
Late work.\ vs.\ Young   & \textbf{\boldmath$+0.0991$} & \scriptsize[$+0.0624$, $+0.1353$] \\
Early work.\ vs.\ Young  & \textbf{\boldmath$+0.0500$} & \scriptsize[$+0.0164$, $+0.0832$] \\
Early work.\ vs.\ Late work.\ & \textbf{\boldmath$-0.0491$} & \scriptsize[$-0.0848$, $-0.0133$] \\
Late work.\ vs.\ Senior  & \textbf{\boldmath$-0.0559$} & \scriptsize[$-0.0944$, $-0.0170$] \\
Early work.\ vs.\ Senior & \textbf{\boldmath$-0.1050$} & \scriptsize[$-0.1447$, $-0.0662$] \\
\bottomrule
\end{tabular}
\caption{Signed lexical warmth on the age axis, $A$ minus $B$, over 330 contexts. Age groups: Young adult (18--24), Early working (25--44), Late working (45--64), Senior (65+). All five components: Table~\ref{tab:app-signed-age-full}.}
\label{tab:signed-age}
\end{table}
\begin{table*}[t]
\centering
\small
\setlength{\tabcolsep}{4pt}
\begin{tabular}{lrrrrr}
\toprule
Contrast ($A$ vs.\ $B$) & PBI & Agency & Certainty & Imperatives ($\lambda I$) & Warmth \\
\midrule
Early work.\ vs.\ Late work.\ & $-0.0664$ & \textbf{\boldmath$-0.1118$} & $+0.0333$ & $+0.0121$ & \textbf{\boldmath$-0.0491$} \\
Early work.\ vs.\ Senior      & $-0.0289$ & \textbf{\boldmath$-0.1053$} & \textbf{\boldmath$+0.0697$} & $+0.0067$ & \textbf{\boldmath$-0.1050$} \\
Early work.\ vs.\ Young       & $-0.0309$ & $-0.0588$ & $+0.0273$ & $+0.0006$ & \textbf{\boldmath$+0.0500$} \\
Late work.\ vs.\ Senior       & $+0.0375$ & $+0.0066$ & $+0.0364$ & $-0.0055$ & \textbf{\boldmath$-0.0559$} \\
Late work.\ vs.\ Young        & $+0.0355$ & $+0.0530$ & $-0.0061$ & $-0.0115$ & \textbf{\boldmath$+0.0991$} \\
Senior vs.\ Young             & $-0.0020$ & $+0.0465$ & $-0.0424$ & $-0.0061$ & \textbf{\boldmath$+0.1550$} \\
\bottomrule
\end{tabular}
\caption{Signed lexical diagnostics for all six age contrasts and all five components, $A$ minus $B$, over 330 contexts with percentile clustered-bootstrap intervals. Bold marks an interval excluding zero. Imperatives are the weighted term $\lambda I(y)$ with $\lambda=0.1$, so PBI $=$ Agency $+$ Certainty $+$ Imperatives.}
\label{tab:app-signed-age-full}
\end{table*}

\section{Objective-Transfer Diagnostic}
\label{app:objective_transfer}
Table~\ref{tab:objective-mismatch} reports the mean absolute PBI gap and the
four-dimension framing score of each PBI-targeted method alongside Original,
under the selection judge and the independent evaluator, Gemma. Every method reduces
the PBI gap while \emph{both} the selection judge and Gemma report increased framing disparity. The paired contrasts
under both independent evaluator families are in Table~\ref{tab:rq3-cross-family}
(Section~\ref{sec:results_objective}).
\begin{table*}[t]
\centering
\small
\setlength{\tabcolsep}{5pt}
\begin{tabular}{llccc}
\toprule
Axis & Method & \shortstack{Mean PBI\\gap} &
\shortstack{FS\\(selection)} & \shortstack{FS\\(Gemma)} \\
\midrule
Gender & Original        & 0.7272 & 1.0470 & 0.7080 \\
       & Lexical edit    & \textbf{0.3088} & 1.1617 & 0.7606 \\
       & Unfiltered Qwen & 0.5654 & 1.1510 & 0.7977 \\
       & Filtered Qwen   & 0.5040 & 1.0818 & 0.7413 \\
\midrule
Age    & Original        & 0.7444 & 1.0778 & 0.8178 \\
       & Lexical edit    & \textbf{0.2984} & 1.2023 & 0.8592 \\
       & Unfiltered Qwen & 0.5850 & 1.1719 & 0.8862 \\
       & Filtered Qwen   & 0.5384 & 1.1102 & 0.8444 \\
\bottomrule
\end{tabular}
\caption{{\small Objective-transfer diagnostic. All methods target PBI and none is selected using either reported framing score. FS (selection) is the equal-weight four-dimension score from the DeepSeek-V3.1 selection judge; FS (Gemma) is the same aggregate from the independent evaluator, Gemma-4-31B-IT. Paired contrasts under both evaluator families: Table~\ref{tab:rq3-cross-family}.}}
\label{tab:objective-mismatch}
\end{table*}

\section{Selective-abstention Diagnostic}
\label{app:reject_only}

Reject-only applies no rewrite. It retains a pair only when its original PBI
gap is at most 0.5, leaving 286/660 gender pairs (43.3\% coverage) and 823/1,980
age rows (41.6\% coverage) (Table \ref{tab:app-reject-only}). On these retained subsets, mean absolute PBI gaps
are 0.0955 for gender and 0.1088 for age, and selection-judge framing scores
are 1.0157 and 1.0279, respectively. Because retention is determined by the
PBI outcome itself, these values describe a selected population rather than a
full-coverage mitigation policy. Reject-only is consequently reported as a
selective-abstention diagnostic and is not ranked directly against rewriting
methods that return an output for every pair.

\begin{table}[h]
\centering
\footnotesize
\setlength{\tabcolsep}{3pt}
\begin{tabular}{lrrrr}
\toprule
Axis & Retained & Coverage & \shortstack{Mean PBI\\gap} & \shortstack{Selection\\judge\\framing} \\
\midrule
Gender & 286/660 & 43.3\% & 0.0955 & 1.0157 \\
Age & 823/1,980 & 41.6\% & 0.1088 & 1.0279 \\
\bottomrule
\end{tabular}
\caption{Reject-only ($d_P\leq0.5$) selective-abstention results. Retention is
determined by the original PBI gap, so these are selected subsets, not
comparable with full-coverage rewriting policies.}
\label{tab:app-reject-only}
\end{table}
\clearpage
\onecolumn

\begin{small}
\setlength{\tabcolsep}{5pt}
\renewcommand{\arraystretch}{1.15}

\begin{longtable}{
    >{\RaggedRight\arraybackslash}p{0.13\textwidth}
    >{\RaggedRight\arraybackslash}p{0.35\textwidth}
    >{\RaggedRight\arraybackslash}p{0.24\textwidth}
    >{\RaggedRight\arraybackslash}p{0.18\textwidth}
}

\caption{
Automatically selected qualitative examples. Examples are selected using
prespecified automated criteria. Negative disparity deltas indicate improvement.
}
\label{tab:qualitative-examples}
\\

\toprule
\textbf{Category \& setting} &
\textbf{Original Pair \& Edited Output} &
\textbf{Automated Metrics} &
\textbf{Interpretation} \\
\midrule
\endfirsthead

\multicolumn{4}{c}{
\tablename~\thetable{} continued
}
\\
\toprule
\textbf{Category and setting} &
\textbf{Original pair and edited output} &
\textbf{Automated metrics} &
\textbf{Interpretation} \\
\midrule
\endhead

\midrule
\multicolumn{4}{r}{Continued on the next page}
\\
\endfoot

\bottomrule
\endlastfoot

\textbf{Independent success}

\medskip
Age

\texttt{adaptive\_single}
&
\textbf{Original A:}
``Midwest sisters, you've seen America grow \& glow! Let's fuel our
nation's future, embrace Pro-Energy. Your wisdom, our power. United, we
shine!''

\medskip
\textbf{Original B:}
``Empower the Heartland, fuel our future! Young women of the Midwest,
answer the call. By embracing pro-energy policies, you're sparking
independence, driving innovation,\ldots''

\medskip
\textbf{Edited A:}
``Midwest sisters, witness America's growth and glow! Let's power our
nation's future with Pro-Energy. Your wisdom fuels our strength. Together,
we shine!''
&
\begin{tabular}[t]{@{}ll@{}}
\(\Delta\) selection: & \(-0.250\)\\
\(\Delta\) Gemma dim.: & \(-0.250\)\\
\(\Delta\) Gemma overall: & \(-1.000\)\\
Cosine: & \(0.911\)\\
Entailment: & \(0.993\)\\
Contradiction: & \(0.001\)\\
Content: & \(1\)\\
\(\Delta\) mean PBI: & \(+0.050\)
\end{tabular}
&
Selection and independent outcomes improve with strong automated
preservation. 
\\

\midrule

\textbf{Judge disagreement}

\medskip
Age

\texttt{adaptive\_single}
&
\textbf{Original A:}
``Midwest women 45--64, let's fuel our future together! Adopting
Pro-Energy habits boosts local economy, lowers utility bills, and empowers
us. Every switch, every LED bulb\ldots''

\medskip
\textbf{Original B:}
``Empower your community, adopt Pro-Energy! As a senior Midwest woman,
you've seen changes. Boost local jobs, strengthen economy, embrace
renewable energy. Your support ca\ldots''

\medskip
\textbf{Edited A:}
``Midwest women, let's fuel our future together! Adopting Pro-Energy habits
boosts our local economy, lowers utility bills, and empowers us. Every
switch, every LED bulb,\ldots''
&
\begin{tabular}[t]{@{}ll@{}}
\(\Delta\) selection: & \(-1.250\)\\
\(\Delta\) Gemma dim.: & \(0.000\)\\
\(\Delta\) Gemma overall: & \(+1.000\)\\
Cosine: & \(0.852\)\\
Entailment: & \(0.003\)\\
Contradiction: & \(0.000\)\\
Content: & \(1\)\\
\(\Delta\) mean PBI: & \(-0.450\)
\end{tabular}
&
The selection score improves, but at least one independent outcome worsens,
illustrating evaluator non-transfer.
\\

\midrule

\textbf{Semantic risk}

\medskip
Age

\texttt{adaptive\_single}
&
\textbf{Original A:}
``Senior Southeastern ladies, you've seen America grow! Let's fuel
tomorrow's dreams by embracing pro-energy policies. It's patriotic to
support independence, jobs, and p\ldots''

\medskip
\textbf{Original B:}
``Empower the Southeast, empower America! As young women, let's lead in
pro-energy adoption. It's not just about us, it's about our community, our
country, and our future\ldots''

\medskip
\textbf{Edited B:}
``Empower the Southeast, empower America! As senior women, let's lead in
pro-energy policies. It's about our community, our country, and our future.
Every small step---effic\ldots''
&
\begin{tabular}[t]{@{}ll@{}}
\(\Delta\) selection: & \(-0.500\)\\
\(\Delta\) Gemma dim.: & \(0.000\)\\
\(\Delta\) Gemma overall: & \(0.000\)\\
Cosine: & \(0.886\)\\
Entailment: & \(0.002\)\\
Contradiction: & \(0.995\)\\
Content: & \(2\)\\
\(\Delta\) mean PBI: & \(-0.550\)
\end{tabular}
&
The rewrite changes the audience from young to senior women, replaces
behavioral adoption with policy support, and removes concrete efficiency
actions. This is observable content drift rather than a human judgment.
\\

\midrule

\textbf{Persuasion amplification}

\medskip
Age

\texttt{both\_sides}
&
\textbf{Original A:}
``Join your Southeast neighbors in leading the nation! As a wise senior,
you've seen changes. Embrace one more: clean energy. It's great for the
grandkids, your bills,\ldots''

\medskip
\textbf{Original B:}
``Join the Southeast's clean-energy revolution! As a young woman, your
actions today shape our tomorrow. Embrace solar, wind, or electric
vehicles. Every switch matters,\ldots''

\medskip
\textbf{Edited A:}
``Join your Southeast neighbors in leading the nation! As a wise senior,
you've seen changes. Embrace one more: clean energy. It's great for the
grandkids, your bills,\ldots''

\medskip
\textbf{Edited B:}
``Join the Southeast's clean-energy revolution! As a wise senior, you've
seen changes. Embrace solar, wind, or electric vehicles. Your choices today
shape a sustainable f\ldots''
&
\begin{tabular}[t]{@{}ll@{}}
\(\Delta\) selection: & \(-0.500\)\\
\(\Delta\) Gemma dim.: & \(-0.500\)\\
\(\Delta\) Gemma overall: & \(-1.000\)\\
Cosine: & \(0.885\)\\
Entailment: & \(0.000\)\\
Contradiction: & \(0.999\)\\
Content: & \(1\)\\
\(\Delta\) mean PBI: & \(+1.050\)
\end{tabular}
&
Independent disparity improves while mean pair-level PBI increases,
indicating equalization through amplification.
\\

\midrule

\textbf{Unnatural output}

\medskip
Age

\texttt{both\_sides}
&
\textbf{Original A:}
``Southwest men 45--64, reduce pollution-related health risks by adopting
clean energy. Transition to solar, wind, or hybrid vehicles to breathe
easier, live longer, and pr\ldots''

\medskip
\textbf{Original B:}
``Southwest seniors, prioritize your health and the planet's. Switch to
clean energy, breathe easier, and reduce pollution-related illnesses.
Invest in a sustainable futu\ldots''

\medskip
\textbf{Edited A:}
``Reduce pollution-related health risks by switching to clean energy.
Choose solar, wind, or hybrid vehicles to breathe easier, live longer, and
protect your heart health\ldots''

\medskip
\textbf{Edited B:}
``\texttt{<<<}Southwest seniors, prioritize your health and the planet's.
Switch to clean energy, breathe easier, and reduce pollution-related
illnesses. Embrace a sustainable futu\ldots''
&
\begin{tabular}[t]{@{}ll@{}}
\(\Delta\) selection: & \(-0.500\)\\
\(\Delta\) Gemma dim.: & \(0.000\)\\
\(\Delta\) Gemma overall: & \(-1.000\)\\
Cosine: & \(0.705\)\\
Entailment: & \(0.003\)\\
Contradiction: & \(0.017\)\\
Content: & \(1\)\\
\(\Delta\) mean PBI: & \(0.000\)
\end{tabular}
&
The retained output contains an observable formatting or repetition artifact
despite passing the automated acceptance rule.
\\

\midrule

\textbf{Age-inconsistent context}

\medskip
Age

\texttt{adaptive\_single}
&
\textbf{Original A:}
``Wild mustangs roam free in the West. Let's unleash clean energy,
gentlemen. Join the stampede toward a sustainable future, reducing carbon
hooves and preserving our ma\ldots''

\medskip
\textbf{Original B:}
``Wild West seniors, let's ride into a sustainable future! Adopt clean
energy, just like a proud eagle soars on wind currents, and help preserve
the American wilderness f\ldots''

\medskip
\textbf{Edited A:}
``Wild mustangs roam free in the West. Let's unleash clean energy,
gentlemen. Join the stampede toward a sustainable future, just like the
eagles soaring high in the west\ldots''
&
\begin{tabular}[t]{@{}ll@{}}
\(\Delta\) selection: & \(-0.500\)\\
\(\Delta\) Gemma dim.: & \(+0.250\)\\
\(\Delta\) Gemma overall: & \(+1.000\)\\
Cosine: & \(0.843\)\\
Entailment: & \(0.002\)\\
Contradiction: & \(0.002\)\\
Content: & \(1\)\\
\(\Delta\) mean PBI: & \(0.000\)
\end{tabular}
&
The policy improves some age contrasts and worsens others within the same
context, meaning that a context-level mean can conceal contrast-level
deterioration. Across six contrasts, the dimension-level counts for
improved/tied/worsened are \(1/4/1\), and the overall counts are \(1/4/1\).
\\

\end{longtable}
\end{small}

\end{document}